\documentclass[times, review, 10pt]{elsarticle}

\usepackage{amssymb}
\usepackage{amsmath}

\usepackage[dvipsnames]{xcolor}
\usepackage{algorithm}
\usepackage{subcaption}
\usepackage{graphicx}
\usepackage{amsfonts}
\usepackage[fleqn]{nccmath}
\usepackage{pifont}
\usepackage{algpseudocode}
\usepackage{hhline}
\usepackage{multirow}
\usepackage{makecell}
\usepackage{comment}
\usepackage{url}
\usepackage{hyperref}

\newenvironment{compactalgorithm}
  {\begin{algorithm} 
   \setlength{\baselineskip}{9pt} 
  }
  {\end{algorithm}}

\newcommand{\etal}{\emph{et al.}}
\newcommand{\cmark}{\ding{51}} 
\newcommand{\xmark}{\ding{55}} 

\def\numImg{{K}}

\def\bboxSetBase{\mathcal{D}}
\def\bboxSet{\bboxSetBase^{3D}}
\def\bboxSetLidar{\widehat{\bboxSetBase}^{3D}}

\def\imgSet{{\mathcal{I}}}
\def\pointCloud{{\mathcal{P}}}
\newcommand{\frustumProposal}[1]{\pointCloud_{#1}}
\def\clsSet{{\Lambda}}
\def\clsSymbol{{\lambda}}
\newcommand{\img}[1]{I_{#1}}
\newcommand{\stereoPair}[1]{(\img{#1}^l, \img{#1}^r)}
\def\bboxSetImages{\mathcal{D}^{2D}}
\newcommand{\bboxSetImgStereo}[1]{\bboxSetBase_{#1}^{2D}}

\def\TrMat{{T}}
\def\ProjMatrix{{P}}
\newcommand{\pMat}[1]{\ProjMatrix_{#1}}
\newcommand{\fMat}[2]{F_{#1#2}}

\newcommand{\cls}[1]{\clsSymbol_{#1}}
\def\yawAngle{{\theta}}
\newcommand{\yaw}[1]{\yawAngle_{#1}}
\newcommand{\bboxGeneric}[1]{\mathbf{b}^{#1}}
\newcommand{\bboxLidar}[1]{\mathbf{B}_{#1}}
\newcommand{\bboxImg}[1]{\mathbf{b}_{#1}}

\newcommand{\real}[1]{\mathbb{R}^{#1}}

\def\Matching{{\mathcal{M}}}

\def\MatchingRecover{{\mathcal{R}}}
\def\MatchingFinal{{\mathcal{A}}}
\newcommand{\MatchingSingle}[2]{\Matching_{#1}^{#2}}
\newcommand{\MatchingSingleCluster}[1]{\widetilde{\Matching_{#1}}}

\def\bboxSetImagesUnmatch{{\mathcal{U}}}
\newcommand{\bboxSetImgUnmatch}[2]{\bboxSetImagesUnmatch_{#1}^{#2}}
\newcommand{\bboxSetImgUnmatchStereo}[1]{\bboxSetImagesUnmatch_{{#1}}}

\def\varSymbol{{x}}

\def\iouThrBBoxMatching{\tau_b}
\def\iouThrDetRec{\tau_d}
\def\minPointFrustum{p_{\min}}
\def\enlargeFactor{e}

\journal{Pattern Recognition}

\begin{document}

\begin{frontmatter}



\title{LCF3D: A Robust and Real-Time Late-Cascade Fusion Framework for 3D Object Detection in Autonomous Driving
\thanks{This paper has been accepted for publication in 
\emph{Pattern Recognition}, 2026. 
The final version is available at 
\href{https://doi.org/10.1016/j.patcog.2026.113046}{https://doi.org/10.1016/j.patcog.2026.113046}.}}



\author[label1]{Carlo Sgaravatti\corref{cor1}}
\ead{carlo.sgaravatti@polimi.it}
\author[label1]{Riccardo Pieroni}
\ead{riccardo.pieroni@polimi.it}
\author[label1]{Matteo Corno}
\ead{matteo.corno@polimi.it}
\author[label1]{Sergio M. Savaresi}
\ead{sergio.savaresi@polimi.it}
\author[label1]{Luca Magri}
\ead{luca.magri@polimi.it}
\author[label1]{Giacomo Boracchi}
\ead{giacomo.boracchi@polimi.it}

\cortext[cor1]{Corresponding Author}

\affiliation[label1]{organization={DEIB- Dipartimento Elettronica, Informazione e Bioingegneria, Politecnico di Milano},
            addressline={Via Ponzio 34/5},
            city={Milan},
            postcode={20133},
            country={Italy}}

\begin{abstract}
Accurately localizing 3D objects like pedestrians, cyclists, and other vehicles is essential in Autonomous Driving. To ensure high detection performance, Autonomous Vehicles complement RGB cameras with LiDAR sensors, but effectively combining these data sources for 3D object detection remains challenging.
We propose LCF3D, a novel sensor fusion framework that combines a 2D object detector on RGB images with a 3D object detector on LiDAR point clouds. By leveraging multimodal fusion principles, we compensate for inaccuracies in the LiDAR object detection network. 
Our solution combines two key principles: (i) \emph{late fusion}, to reduce LiDAR False Positives by matching LiDAR 3D detections with RGB 2D detections and filtering out unmatched LiDAR detections; and (ii) \emph{cascade fusion}, to recover missed objects from LiDAR by generating new 3D frustum proposals corresponding to unmatched RGB detections. Experiments show that LCF3D is beneficial for domain generalization, as it turns out to be successful in handling different sensor configurations between training and testing domains.
LCF3D achieves significant improvements over LiDAR-based methods, particularly for challenging categories like pedestrians and cyclists in the KITTI dataset, as well as motorcycles and bicycles in nuScenes. Code can be downloaded from: \url{https://github.com/CarloSgaravatti/LCF3D}.
\end{abstract}



\begin{keyword}


Deep Learning \sep 3D Object Detection \sep Multimodal \sep Autonomous Vehicles
\end{keyword}

\end{frontmatter}




\section{Introduction} \label{sec:introduction}

In Autonomous Driving (AD), a key requirement for safe navigation is the accurate detection of small and distant objects, such as pedestrians, cyclists, and other vulnerable road users. This task is particularly challenging because autonomous vehicles must detect such objects in real time, often by combining multiple sensing modalities, each with its own limitations and latency constraints. To achieve this goal, Autonomous Vehicles (AVs) are typically equipped with complementary sensors such as LiDAR scanners and RGB cameras. LiDAR sensors provide accurate geometric information but return extremely sparse measurements for distant objects, making small targets difficult to detect and often underrepresented in AD datasets \cite{kitti_dataset,nuscenes2019}. Conversely, RGB images offer dense semantic information that improves the recognition of small and distant objects \cite{survey_mm_autonomous_drive}, but lack explicit depth cues, which limits precise 3D localization \cite{qian20223d}.
Combining both sensing modalities is crucial to achieve both accurate 3D localization and robust semantic understanding. However, effectively fusing LiDAR and RGB data without introducing substantial computational overheads that hinder real-time performance (LiDAR typically operates at 10-20 Hz) remains challenging. 

Several \emph{fusion} strategies have been proposed in AD to combine LiDAR and RGB information, differing in the stage where the two modalities are fused.
\emph{Early fusion} \cite{mvx-net,point-painting} injects RGB information directly into the point cloud before it is processed by the LiDAR-based detector. A particular case of this approach is \emph{Cascade fusion} \cite{frustum-pointnet}, which first detects 2D objects in RGB images and then generates 3D frustums from the corresponding regions to guide LiDAR processing. Although effective in reducing the 3D search space, these methods are computationally expensive, as the two modalities cannot be processed in parallel and each frustum requires separate inference, making them unsuitable for real-time applications.
\emph{Intermediate fusion} approaches \cite{li2023logonet,liu2023bevfusion} merge the feature representations extracted from both modalities within a single end-to-end deep network. While they exploit rich cross-modal interactions, the joint training and feature alignment significantly increase their computational cost, preventing real-time operation.
Finally, \emph{late fusion} methods \cite{clocs_late_fusion,long_tailed_late_fusion} process LiDAR and RGB data independently and combine their predictions to suppress LiDAR False Positives (FPs). The two branches can be processed in parallel, but objects missed by the LiDAR detector, i.e. False Negatives (FNs), cannot be recovered, thus small and distant targets often remain undetected \cite{survey_mm_autonomous_drive}.

Beyond computational challenges, a further critical issue when performing 3D object detection in real-world AD scenarios is \emph{domain shift}, which causes performance degradation when models are deployed in an environment different from the training one. Such shifts arise naturally in AD due to variations in sensor configurations, weather conditions, or geographic settings. For instance, LiDAR detectors are sensitive to changes in beam density, while RGB-based models depend on camera intrinsics and optical properties, often leading to reduced accuracy when transferred across setups \cite{wang2023towards}. Since collecting and annotating large multimodal datasets for every possible configuration is very expensive, models must be designed to generalise well across domains, which is preferred over relying on exhaustive re-training \cite{wozniak2024uada3d}.
In particular, we address domain generalization, aiming for models that preserve accuracy across different environments and sensor setups without requiring model fine-tuning and access to target-domain data. In our experiments, we assess domain generalization performances of LCF3D through variations in the sensing equipment, which is a practical proxy that reflects real-world changes such as hardware differences or weather conditions that commonly affect 3D object detection.

We propose \emph{Late Cascade Fusion 3D} (LCF3D), a novel hybrid fusion framework that succesfully integrates principles from both late and cascade fusion paradigms, combining the outputs of a 2D RGB object detector and a 3D LiDAR-based object detector. In practice, LCF3D is designed to address two critical limitations of LiDAR-based 3D detection: the presence of FPs and the failure to detect small or distant objects due to Point Cloud sparsity. To this end, LCF3D introduces three key components: i) a \emph{Bounding Box Matching} module for filtering FPs and increasing the precision of the detector, ii) a \emph{Detection Recovery} module for retrieving missed objects, increasing detection recall, and iii) a \emph{Semantic Fusion} module for resolving label inconsistencies by favouring the semantic predictions of the RGB detector.

More in detail, in the Bounding Box Matching module, we project 3D bounding boxes from the LiDAR branch onto the image plane and match them with 2D detections by an optimization procedure based on Intersection over Union (IoU). LiDAR detections that do not correspond to any RGB detection are considered FPs and removed. Conversely, we consider 2D detections without a corresponding 3D bounding box as potential FNs from the LiDAR branch. Thus, in the Detection Recovery module, we backproject unmatched 2D boxes into 3D frustums where we apply an ad-hoc 3D localization model to recover the missing objects. The Semantic Fusion module finally resolves any label inconsistencies between detections from the two modalities, assigning the final label based on the class predicted by the RGB detector, which we consider more semantically reliable. Our framework, described in Section~\ref{sec:method}, can work on AVs equipped with both monocular (Section \ref{sec:method_single_view}) and stereo images (Section \ref{sec:stereo-view-case}).

Moreover, LCF3D includes lightweight post-processing modules that avoids significant computational overhead associated with early fusion methods, since it can conveniently run detection networks on the RGB and LiDAR data in parallel. Indeed, we exploit cascade fusion principles only for recovering missed objects from the LiDAR branch, while avoiding the substantial computational burden that characterizes traditional cascade-fusion methods that need to process all the RGB detections using frustums. Moreover, the architecture is independent from the underlying 2D and 3D detectors used, without requiring the joint training of the two detectors, enabling the use of off-the-shelf detectors (potentially trained on different single-modal datasets), resulting in great flexibility. Notably, our experiments demonstrates that LCF3D better generalizes to different domains as the 2D RGB Object Detector mitigates the impact of domain shifts due to changes in the LiDAR sensor.

In Section \ref{sec:exp}, we test LCF3D on both KITTI \cite{kitti_dataset} and nuScenes \cite{nuscenes2019} datasets, showing superior performance than LiDAR-based object detectors, especially on imbalanced classes and small objects such as \emph{Pedestrians} and \emph{Cyclists} on KITTI and \emph{Bicycles} and \emph{Motorcycles} on nuScenes. Additionally, by testing models trained on KITTI and on nuScenes, and vice versa, we show that LCF3D generalizes better than other early and intermediate fusion approaches.

Our main contributions can be summarized as follows:
\begin{enumerate}
    \item We propose a novel hybrid LiDAR-RGB fusion method combining late and cascade fusion, that can work with both stereo images and monocular images.
    \item We reduce LiDAR False Positive detections thanks to a novel Bounding Box Matching module applied to clusters of overlapping 3D bounding boxes referred to the same objects detected by the LiDAR branch.
    \item We recover objects missed from LiDAR branch in our Detection Recovery module, which, in the stereo-view setting, leverages epipolar geometry principles to match pairs of 2D detections from different views.
    \item We analyze the effect of domain shifts for 3D LiDAR detectors and study how fusion with 2D RGB detectors can help mitigate them.
\end{enumerate}
This work extends our workshop paper \cite{me}, by: \emph{i)} handling single-view RGB cameras, \emph{ii)} enhancing our Bounding Box Matching module by clustering bounding boxes to improve the matching of 3D detections with 2D detections, \emph{iii)} using Instance Segmentation masks instead of simple 2D bounding boxes in the Detection Recovery module, to further reduce the computational overhead of cascade fusion by selecting fewer points in each frustum, and \emph{iv)} presenting an extended experimental validation to assess domain generalization performance.

\section{Related Work}\label{sec:related_work}

\subsection{Multimodal 3D Object Detection}

Based on how the RGB Images and LiDAR Point Clouds are fused, multimodal approaches can be divided into three categories \cite{mao20233d}: early, intermediate and late fusion.

\subsubsection{Early Fusion}
In Early Fusion approaches, RGB information is integrated into the point cloud before being processed by a LiDAR-based detector.
MVX-Net \cite{mvx-net} projects 3D voxels onto images and concatenates the corresponding pixel features to voxels, while PointPainting \cite{point-painting} attaches semantic labels from image segmentation to each 3D point. These methods suffer from feature blurring, since identical pixel features are often assigned to multiple neighboring points. Other solutions, such as PVConvNet \cite{pvconvnet-pr} and VirConv \cite{virconv}, use depth completion to generate dense pseudo point clouds, but this dramatically increases computational cost, hindering real-time processing.\\
A special case of Early Fusion is Cascade Fusion \cite{survey_mm_autonomous_drive}, where 2D detections from RGB images define frustums that constrain the LiDAR search space \cite{frustum-pointnet}. Although this strategy improves localization, it is computationally expensive because each frustum requires separate 3D inference. Variants like Frustum PointPillars \cite{frustum_pointpillars} mitigate this cost by processing all frustums jointly, but still struggle with multi-class detection and real-time constraints.
In our framework, we adopt a more efficient variant of cascade fusion, applying frustum-based processing only to RGB detections that are not matched with those from the LiDAR branch, thus significantly reducing the computational overhead.

\subsubsection{Intermediate Fusion}

Intermediate Fusion solutions combine the features extracted from the single-modal backbones of 2D and 3D networks in an intermediate layer of an end-to-end trainable multimodal network. RoI-based fusion methods \cite{mv3d} fuse features at a Region of Interest (RoI) level, after finding an initial set of 3D proposals, e.g. from the Bird's Eye View (BEV). These solutions cannot capture cross-modality interactions in the early stages of the network \cite{survey_mm_autonomous_drive}. Differently, recent solutions like BEVFusion \cite{liu2023bevfusion} build a unified representation between LiDAR and RGB images on the BEV, which is more computationally efficient to process than voxels. However, this method is still computationally infeasible for real-time processing as it works at 8.6 FPS on nuScenes, while modern LiDAR sensors work at least at 10 Hz \cite{qian20223d}.

\subsubsection{Late Fusion}

Late fusion approaches employ two parallel Deep Learning branches for the two modalities, namely 3D Object Detection on the LiDAR branch and 2D Object Detection on the RGB branch, and combine their predictions in a fusion network. Typically, late fusion solutions aims at removing FP detections of the 3D LiDAR Object Detection network, by leveraging geometric and semantic consistency among detections from different modalities \cite{clocs_late_fusion}, by adopting an additional 3D object detectors from RGB images \cite{towards_long_tailed}, or by projection of the 3D detections on the image plane \cite{long_tailed_late_fusion}. 
Unfortunately, none of these methods address the recovery of objects missed from LiDAR detectors, which frequently occur for small, distant objects such as pedestrians and cyclists.
Our method, in contrast, is designed to recover accurate 3D detections of all the objects returned by the 2D object detection networks.

\subsection{Domain Adaptation and Generalization}

LiDAR-based detectors are highly sensitive to domain shifts, such as changes in point-cloud sparsity across different sensors or vehicle sizes across different geographic regions. Unsupervised Domain Adaptation (UDA) methods address domain shifts by adapting models from a labeled source domain to an unlabeled target one. Self-training approaches like ST3D \cite{yang2021st3d} use pseudo-labels for fine-tuning, but Zhang \etal \cite{zhang2024revisiting} show that psuedo labels mainly transfer knowledge to the target domain while degrading performance on the source.
In this work, we instead pursue \emph{Domain Generalization} (DG), which aims at training models that are robust across \emph{all} domains without relying on target-domain data. DG remains underexplored for multimodal detection, and Zhang \etal \cite{zhang2024revisiting} report that multimodal models can generalize even worse than LiDAR-only ones. More recently, Hegde \etal \cite{hegde2024multimodal} proposed a supervised contrastive-learning framework to improve multimodal invariance, but it requires extensive cross-domain training.
Our approach achieves domain generalization by an ad-hoc fusion procedure with 2D RGB detectors, which tend to generalize better across varying conditions.

\section{Problem Formulation}

We address a multi-modal multi-class 3D Object Detection problem, where our inputs are a set of $\numImg$ monocular or stereo images $\imgSet$ and a Point Cloud $\pointCloud$. We assume all the sensors are synchronized, i.e. $\imgSet$ and $\pointCloud$ are acquired in the same time frame. More in detail, $\pointCloud$ contains $N$ points $\pointCloud = \{p_1, p_2, ...,  p_N\}$, where $p_j = (x_j, y_j, z_j, r_j)^T \in \real{4}$ and $(x_j, y_j, z_j)$ is the position of the point $p_j$, while $r_j$ is the reflectance measured by LiDAR at that point. $\pointCloud$ is expressed in LiDAR coordinates, with $\TrMat \in \real{4 \times 4}$ being the known transformation matrix from LiDAR to camera coordinates, which are in the coordinate system of a reference camera in $\imgSet$.

A Multimodal 3D Object Detection solution process $\imgSet$ and $\pointCloud$ to return a set of 3D bounding boxes $\bboxSet$ surrounding each object in the 3D space:
\begin{equation} \label{eq:problem_form}
(\imgSet, \pointCloud) \longmapsto \bboxSet = \{(\bboxLidar{p}, s_p, \cls{p})  | \bboxLidar{p} \in \real{7}, s_p \in [0, 1], \cls{p} \in \clsSet, p = 1, \dots, P \},
\end{equation}
where $\bboxLidar{p} = (x_p, y_p, z_p, l_p, h_p, w_p, \yaw{p})^T$ contains the 3D coordinates $(x_p, y_p, z_p)$ of the center of the object, the dimensions $(l_p, h_p, w_p)$ of the bounding box, and the yaw angle $\yaw{p} \in [0, 2\pi]$, $s_p \in [0,1]$ denotes the detection confidence score, $\cls{p}$ is the estimated label from the set of classes $\clsSet$, and $P$ is the number of detections.\\
We consider two possible RGB camera configurations, corresponding to two different setups for $\imgSet$:

\textbf{Single-View Cameras}. In this setting, the AV is equipped with $\numImg$ cameras without overlapping fields of view, and we denote the images as $\imgSet = \{\img{1}, ..., \img{\numImg}\}$, $\img{i} \in \real{W_i \times H_i \times 3}$. We assume to know for each image $\img{i}$ the camera matrix $\pMat{i} \in \real{3 \times 4}$, which projects any 3D point in the coordinate system of the image plane. 

\textbf{Stereo-View Cameras}. In this setting, the AV is equipped with $\numImg$ pairs of stereo-cameras, each providing two different views of the same scene. In this case, $\imgSet$ is a set of (left, right) stereo paired images: $\imgSet = \{\stereoPair{1}, ..., \stereoPair{\numImg}\}$, where $\img{i}^l \in \real{W_i^l \times H_i^l \times 3}$ and $\img{i}^r \in \real{W_i^r \times H_i^r \times 3}$ correspond to the left ($l$) and right ($r$) cameras. Each image $\img{i}^q$, with $q \in \{l,r\}$, is acquired with its own camera matrix $\pMat{i}^q \in \real{3 \times 4}$.

\section{Our Method: LCF3D}
\label{sec:method}
\begin{figure}[t]
\centering
\includegraphics[width=0.9\textwidth]{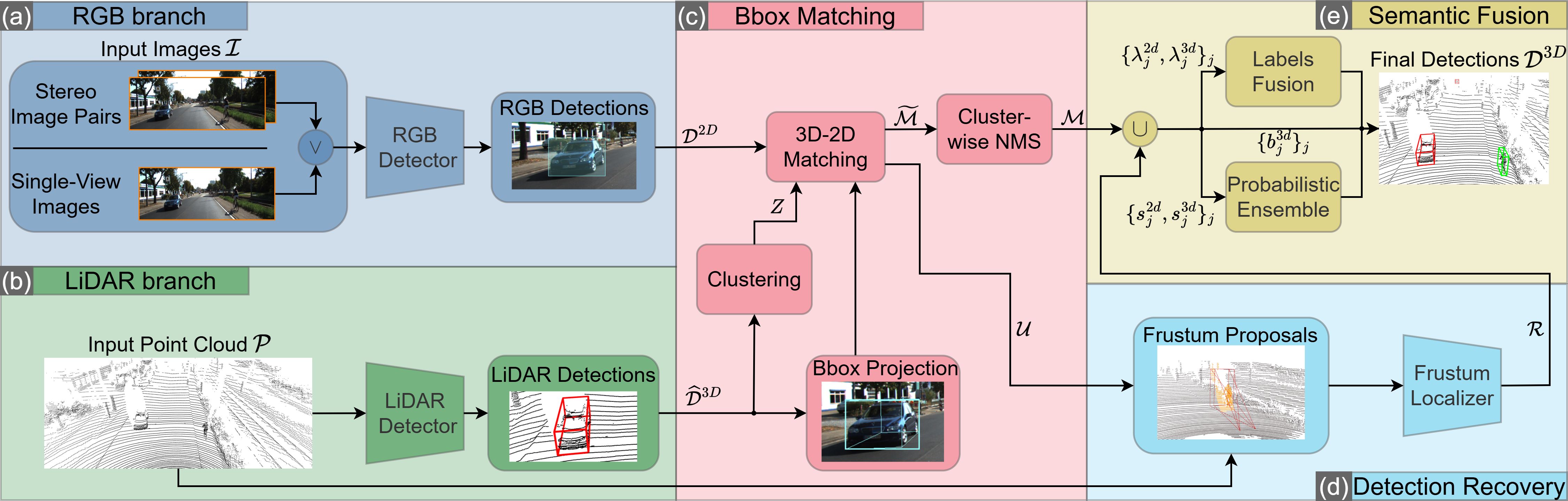}
\caption{
LCF3D consists of two parallel branches and three sequential steps. The RGB branch (a) produces 2D detections $\bboxSetImages$,  the LiDAR branch (b) generates 3D detections $\bboxSetLidar$ from the point cloud $\pointCloud$. In step (c), 3D detections are projected and matched with 2D ones ($\Matching$). Unmatched RGB detections $\bboxSetImagesUnmatch$ are processed in step (d) by the Detection Recovery module, which uses Frustum Proposals and a Frustum Localizer to recover missed LiDAR detections ($\MatchingRecover$). Step (e) employs Semantic Fusion to enforce consistency between LiDAR and RGB branches.
}
\label{fig:architecture}
\end{figure}

At a high level, LCF3D is composed of the 5 modules illustrated in Figure \ref{fig:architecture}: (a) RGB branch, (b) LiDAR branch, (c) Bounding Box Matching, (d) Detection Recovery and (e) Semantic Fusion. The RGB branch takes as input a set $\imgSet$ of single-view or stereo-view images and runs a 2D Object Detection model predicting a set of 2D bounding boxes $\bboxSetImages$. In parallel, the LiDAR branch processes the input Point Cloud $\pointCloud$ to produce a preliminary set of 3D bounding boxes $\bboxSetLidar$. The Bounding Box Matching module performs late fusion to filter out FP detections from the LiDAR branch by matching detections from $\bboxSetImages$ and $\bboxSetLidar$, producing a set $\Matching$ of paired 3D and 2D bounding boxes. Instead, the Detection Recovery step processes all the RGB detections $\bboxSetImgUnmatch{}{}$ that are not matched in (c) to recover FNs from the LiDAR branch (b), producing as output a new set $\MatchingRecover$ of 3D detections associated with unmatched 2D bounding boxes. Finally, Semantic Fusion enforces consistency in predicted labels between the matched detections $\Matching$ in (c) and the new 3D detections $\MatchingRecover$ from (d). The principle underpinning LCF3D is that the RGB modality is better for finding small and distant objects \cite{survey_mm_autonomous_drive}. Thus, we expect the RGB branch to have a higher recall than the LiDAR one for those objects, and rely on cascade fusion in the Detection Recovery for objects missed by the LiDAR branch. This higher recall does not necessarily come at the cost of lower precision: RGB images provide higher spatial resolution and richer appearance cues (texture, color, shape), which makes such objects easier to localize reliably than in sparse LiDAR point clouds \cite{survey_mm_autonomous_drive}. Moreover, the Bounding Box Matching module exploits the fact that RGB detectors are expected to have a high precision, since RGB images possess richer semantics. Please note that our method is not constrained to a particular model in the LiDAR and RGB branches, enabling the usage of any state-of-the-art object detection models (discussed in Section \ref{sec:implementation-details}).

\subsection{The Single View case}
\label{sec:method_single_view}
Fot the sake of illustration, we will outline the method for the single-view setup. We will then discuss the extension to the stereo-view case in Section \ref{sec:stereo-view-case}. For the notation sake, we will also illustrate our method assuming a single -image is processed ($\numImg = 1$); the formulation naturally extends to multiple, non overlapping, images ($\numImg > 1$) by repeating all the steps for each input image.

\subsubsection{RGB branch}

In the RGB branch, a 2D Object Detection network process the input image to produce 2D bounding boxes. Any 2D Object Detection model can be used in this branch: single-stage object detectors are preferred when fast inference is needed, but they usually perform worse than two-stage detectors like Faster RCNN \cite{faster_rcnn}. Since in the AD field objects can have very different depths, a Feature Pyramid Network (FPN) is usually employed as it produces features at different scales, which is useful to detect objects having very different 2D dimensions. Optionally, we can also adopt an instance segmentation network to predict masks within each box to further reduce the search space for Detection Recovery, as discussed in Section \ref{sec:det-recovery-single}.

We denote with $\bboxSetImages = \{(\bboxImg{m}, s_m, \cls{m})\}_{m=1}^{M}$ the set of 2D detections in the image $\img{}$, where $\bboxImg{m}$ is a 2D bounding box described by 4 coordinates $(x_{min}, y_{min}, x_{max}, y_{max})$, $s_m \in [0, 1]$ is the confidence score, $\cls{m} \in \clsSet$ is the semantic class and $M$ is the number of detections. Since we are interested in 3D bounding boxes, we combine $\bboxSetImages$ with detections from a 3D LiDAR Object Detection network.

\subsubsection{LiDAR branch}

The LiDAR branch is a 3D Object Detection network processing Point Clouds $\pointCloud$, which produces an initial set $\bboxSetLidar$ of 3D bounding boxes, as in \eqref{eq:problem_form}, that are further processed by other modules. 
Point-based methods \cite{point-rcnn}, which extract features directly from LiDAR points, are typically too slow for real-time applications. Instead, we follow a Point Cloud discretization approach in either voxels or BEV. Specifically, voxel-based methods \cite{lgnet-pr,hrnet-pr,qian2022badet-pr} encode points into voxels and use 3D CNNs for feature extraction, offering faster inference with comparable performance. BEV-based methods \cite{pointpillars,zhu2020ssn}, instead, project the Point Cloud into the Bird's Eye View (BEV) and use 2D CNNs, enabling faster inference at the cost of additional 2D discretization.

Due to the sparsity and occlusions that affect the Point Cloud, detections in $\bboxSetLidar$ might either miss some relevant object in the scene. Therefore, we remove the Non Maximum Suppression (NMS) step in the LiDAR branch and use a low confidence score threshold in the LiDAR detector. These changes improve the recall of the LiDAR network at the cost of some 3D FPs. We compensate for these by late fusion with RGB detections in the Bounding Box Matching module.

\subsubsection{Bounding Box Matching} \label{sec:bbox-matching-high-level}

\begin{figure}[t]
    \centering
    \includegraphics[width=0.9\linewidth]{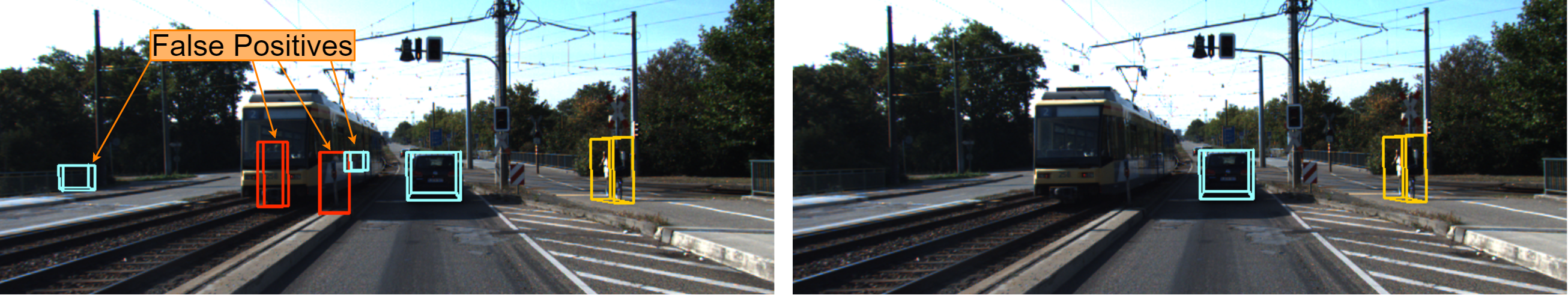}
    \caption{Comparison between the LiDAR branch output (left) and the Bbox Matching module output (right), which removes FPs. Only the highest-confidence bounding box per cluster is shown on the left.
    }
    \label{fig:bbox_match}
\end{figure}

The goal of the Bounding Box Matching module is to remove FP detections from the LiDAR branch by we associating each LiDAR 3D detection in $\bboxSetLidar$ to a 2D detection of $\bboxSetImages$, as shown in Figure \ref{fig:bbox_match}. Projecting 3D detections in the image plane results in information loss (e.g., the depth or the 3D orientation of the bounding box), and the projected 3D detections might not match at best a 2D detection. Therefore, we need to be cautious before discarding a 3D detection yielding low IoU with 2D detections. To this purpose, we replace the NMS of the 3D network with an ad-hoc procedure where neighbouring 3D bounding boxes are first clustered and then matched at cluster-level to 2D detections. This safely discards 3D detections whose corresponding cluster does not overlap with any 2D detection. By keeping a high threshold for both the confidence score and NMS in the 2D Object Detection network, we ensure that clusters of 3D bounding boxes are matched only with relevant 2D detections.

\begin{figure}[t]
\centering
\includegraphics[width=0.9\textwidth]{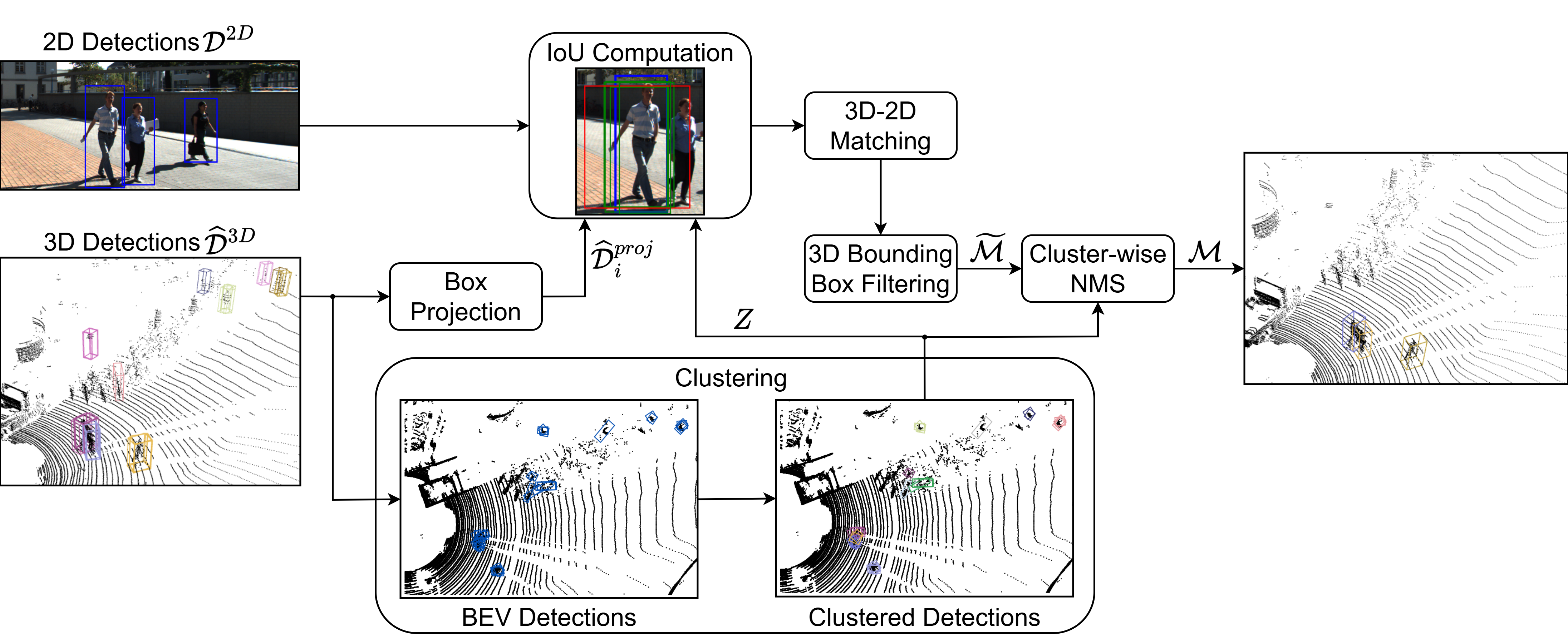}
\caption{
High-level overview of the Bounding Box Matching for one image. First, 3D boxes are clustered in BEV, projected onto the image, and matched with 2D detections using IoU.
Finally, each 3D cluster is matched with the 2D detections using an optimization problem based on the IoU and the bounding box with the highest score inside each matched cluster is selected (Cluster-wise NMS).}
\label{fig:late-fusion}
\end{figure}

\begin{figure}[t]
    \centering
    \includegraphics[width=0.9\linewidth]{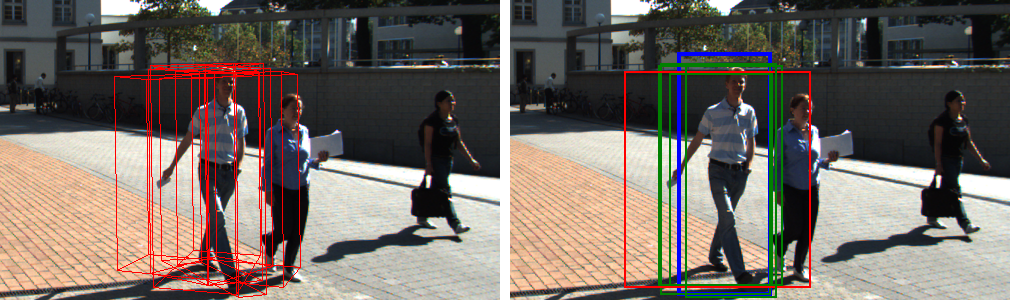}
    \caption{
    (Left) Projection of all the 3D bounding box of a cluster $Z_c$ onto the image plane. (Right) Comparison with RGB 2D detections (blue): green boxes have IoU $> 0.5$, red boxes are below the threshold. The maximum IoU in the cluster is used to resolve conflicts.
    }
    \label{fig:cluster-iou}
\end{figure}

The core components of the Bounding Box Matching module are illustrated in Figure \ref{fig:late-fusion}. When removing the NMS from the 3D Object Detection network, the LiDAR branch returns several 3D detections for a single 3D object (see Figure \ref{fig:cluster-iou} left). Thus, we first \emph{cluster} the 3D detections and then project all the 3D bounding boxes on the image plane, where we solve a \emph{matching} problem to pair 3D clusters with 2D bounding boxes. Unmatched clusters are removed, while matched clusters undergo a \emph{cluster-wise NMS} step to retrieve the highest-confidence 3D bounding box within the cluster.

In practice, rather than performing the clustering directly in 3D, we found it convenient to work on the BEV. We define a cluster of 3D bounding boxes as a subset of $\bboxSetLidar$ having  a high mutual IoU on the BEV. Clusters of 3D bounding boxes are hence identified as maximal cliques in a graph where the nodes are the 3D bounding boxes and an edges connect two bounding boxes when their 2D IoU on the BEV $IoU_{BEV}$ is higher than a threshold $\tau_z$.
More formally, denoting $Z = \{Z_c | c = 1, \dots, C\}$ the set of clusters of 3D detections, a cluster $Z_c$ is identified as:
\begin{equation}\label{eq:cluster_def}
    \forall \bboxLidar{i}, \bboxLidar{j} \in \bboxSetLidar: (\bboxLidar{i} \in Z_c \land \bboxLidar{j} \in Z_c) \iff IoU_{BEV}(\bboxLidar{i}, \bboxLidar{j}) > \tau_z.
\end{equation}


Then, we assign a cluster $Z_c$ to a 2D detection $\bboxImg{m} \in \bboxSetImages$ as follows. We first project the 8 corners $\{h_1, \dots,  h_8\}$ of each 3D bounding box $\bboxLidar{} \in Z_c$ into the image plane using the projection matrix $\pMat{i}$. 
From the projected corners $\widetilde{h}_j = \pMat{i} \TrMat h_j$, we extract the minimum axis-aligned bounding box $\bboxGeneric{proj}_i$  enclosing all of them. We define the 2D IoU between a cluster $Z_c$ and a 2D bounding box $\bboxImg{m} \in \bboxSetImages$ as the maximum IoU of the projected bounding boxes of the cluster with $\bboxImg{m}$, as illustrated in Figure \ref{fig:cluster-iou} right:
\begin{equation}\label{eq:iou_cluster}
    IoU(Z_c, \bboxImg{m}) = \max_{\bboxLidar{i} \in Z_c} IoU_{2d}(\bboxGeneric{proj}_i, \bboxImg{m}).
\end{equation}
 
To compute a matching $\MatchingSingleCluster{}$ between clusters in $Z$ and bounding boxes in $\bboxSetImages$, we solve the following optimization problem using the Jonker-Volgenant algorithm \cite{lsap-alg}:

\begin{subequations}\label{eq:linear_sum_assign}
\noindent
\begin{minipage}[b]{0.07\textwidth}
    $\max_{\varSymbol}$
\end{minipage}
\begin{minipage}[b]{0.425\textwidth}
    \begin{fleqn}
    \begin{equation}
    \sum\limits_{c,m} IoU(Z_c, \bboxImg{m}) \varSymbol_{c,m}, \tag{\ref{eq:linear_sum_assign}}
    \end{equation}
    \end{fleqn}
\end{minipage}
\begin{minipage}[b]{0.06\textwidth}
    \hfill \text{s.t.}
\end{minipage}
\begin{minipage}[b]{0.425\textwidth}
    \begin{fleqn}
    \begin{equation}
    \sum\limits_m \varSymbol_{c,m} \leq 1, \hspace{0.5em} \forall c \in \{1, \dots, C\}, \label{eq:linear_sum_assign:constr1}
    \end{equation}
    \end{fleqn}
\end{minipage}\\
\noindent
\begin{minipage}[b]{0.07\textwidth}
\hfill
\end{minipage}
\begin{minipage}[b]{0.425\textwidth}
    \begin{fleqn}
    \begin{equation}
    \sum\limits_c \varSymbol_{c,m} \leq 1, \hspace{0.5em} \forall m \in \{1, \dots, M\}, \label{eq:linear_sum_assign:constr2}
    \end{equation}
    \end{fleqn}
\end{minipage}
\begin{minipage}[b]{0.06\textwidth}
\hfill
\end{minipage}
\begin{minipage}[b]{0.425\textwidth}
    \begin{fleqn}
    \begin{equation}
    \sum\limits_{c,m} \varSymbol_{c,m} \ge \min\{C, M\}, \label{eq:linear_sum_assign:constr3}
    \end{equation}
    \end{fleqn}
\end{minipage}
\end{subequations}
where $\varSymbol_{c,m} \in \{0,1\}$ denotes if cluster $Z_c$ and 2D detection $\bboxImg{m}$ are matched or not. 
The first two constraints ensure each cluster matches at most one RGB detection, and vice versa, while the last maximizes total assignments.
We keep in $\MatchingSingleCluster{}$ only the matched clusters with an IoU higher than a threshold $\iouThrBBoxMatching$, and for each matched cluster, we keep only the 3D bounding box with the highest confidence score (\emph{cluster-wise NMS}), yielding the set of matches $\MatchingSingle{}{}$. 
The set of unmatched RGB detections, denoted by $\bboxSetImgUnmatch{}{}$, will be processed in the following Detection Recovery module (Section \ref{sec:det-recovery-single}).

Alternative strategies exist to match RGB and LiDAR detections, such as lifting 2D boxes to BEV \cite{long_tailed_late_fusion} or using a 3D detector in the RGB branch \cite{towards_long_tailed}. These approaches generally underperform compared to image-plane matching \cite{long_tailed_late_fusion}, and our experiments (Section \ref{sec:ablation-bbox-match}) confirm the benefit of postponing NMS until after 3D–2D matching.

\subsubsection{Detection Recovery} \label{sec:det-recovery-single}

The Detection Recovery module recovers FNs from the LiDAR branch, i.e. missed 3D objects, processing the set of unmatched 2D bounding boxes for each image $\bboxSetImagesUnmatch \subseteq  \bboxSetImages$, where $\bboxSetImgUnmatch{}{} \subseteq \bboxSetImages$. We recover the corresponding missed 3D detections by leveraging single-view geometry principles, returning a set $\MatchingRecover$ of pairs of RGB-LiDAR detections:
\begin{equation}
    \MatchingRecover := \{(\bboxLidar{j}, s_j^{3d}, \cls{j}^{3d}, \bboxImg{j}, s_{j}^{2d}, \cls{j}^{2d})\},
\end{equation}
where $(\bboxLidar{j}, s_j^{3d}, \cls{j}^{3d})$ are the new 3D detections recovered and $(\bboxImg{j}, s_j^{2d}, \cls{j}^{2d}) \in \bboxSetImages$ is the corresponding 2D detection.

\begin{figure}[t]
\centering
\includegraphics[width=0.9\textwidth]{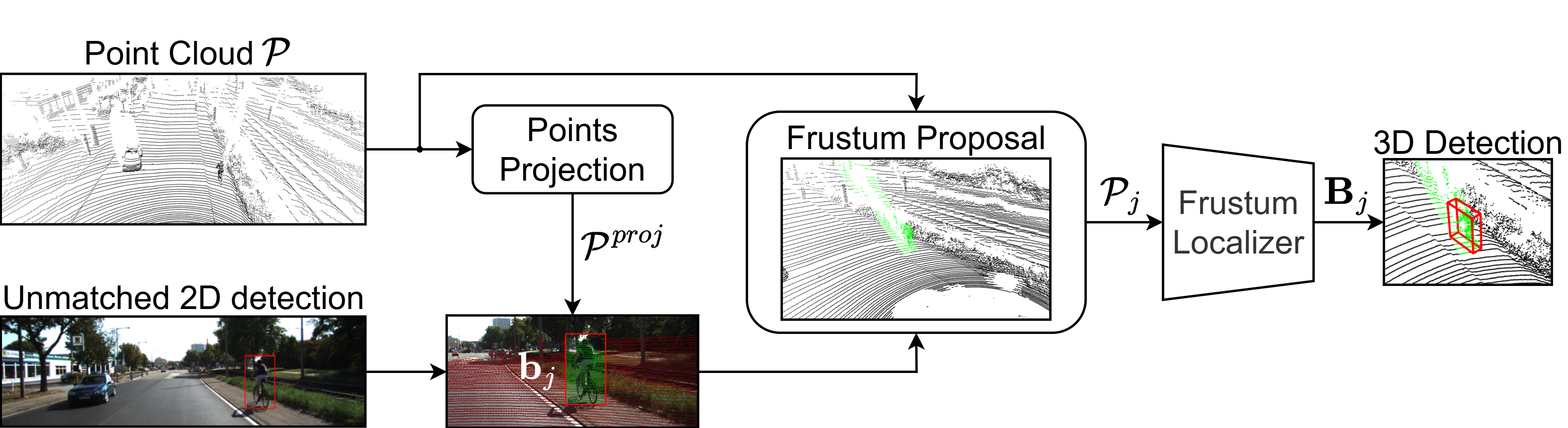}
\caption{
In the single-view scenario, the Detection Recovery module processes an unmatched 2D RGB bounding box $\bboxImg{j}$. A Frustum Proposal $\frustumProposal{j}$ is generated by projecting the 3D points of the input Point Cloud $\pointCloud$ into the image and selecting points within $\bboxImg{j}$. The Frustum Localizer then extracts a 3D bounding box $\bboxLidar{j}$ from the frustum, which inherits the semantic label of the 2D detection.
}
\label{fig:detection-recovery-architecture}
\end{figure}

\begin{figure}
    \centering
    \subfloat[2D detection\label{fig:frustums:masks}]{
        \includegraphics[width=0.2\linewidth]{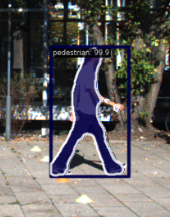}
    }
    \subfloat[Bbox\label{fig:frustums:bbox_frustum}]{
        \includegraphics[width=0.2\linewidth]{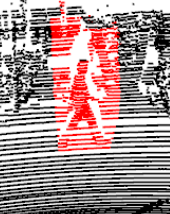}
    }
    \subfloat[Mask\label{fig:frustums:mask_frustum}]{
        \includegraphics[width=0.2\linewidth]{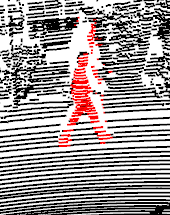}
    }
    \subfloat[Bbox+Mask\label{fig:frustums:mask_frustum_one_hot}]{
        \includegraphics[width=0.2\linewidth]{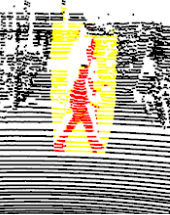}
    }
    \caption{
    (a) Output of an instance segmentation network. (b) Frustum Proposal from a 2D bounding box (red points). (c) Frustum Proposal from a 2D instance mask. (d) Frustum Proposal from a 2D bounding box with mask channel (yellow = 0, red = 1).
    }
    \label{fig:frustums}
\end{figure}

Figure \ref{fig:detection-recovery-architecture} illustrates the Detection Recovery module. We extract from each unmatched 2D bounding box $\bboxImg{j} \in \bboxSetImgUnmatch{}{}$ a Frustum Proposal $\frustumProposal{j}$, i.e. the set of 3D points contained into the 3D frustum of that 2D bounding box. This is illustrated in Figure \ref{fig:frustums:bbox_frustum}, where the frustums are in red and correspond to the set of 3D points that are projected inside the 2D bounding box in Figure \ref{fig:frustums:masks}. Note that, before computing the Frustum Proposal, we slightly increase the dimensions of the 2D bounding boxes by an enlargement factor $\enlargeFactor$ for both width and height, keeping the centers of the bounding boxes fixed. Moreover, we add to each point $p \in \frustumProposal{j}$, projected in the image as $p^{proj} = (x^{proj}, y^{proj})$, the Gaussian mask proposed in Frustum PointPillars \cite{frustum_pointpillars}, computed as:
\begin{equation}
    G(p^{proj}) = \exp \left( -\frac{(x^{proj} - x_0)^2}{2w^2} - \frac{(y^{proj} - y_0)^2}{2h^2} \right),
\end{equation}
where $(x_0, y_0)$ is the center of $\bboxImg{j}$ and $(w, h)$ are the width and the height. We filter out Frustum Proposals that contain less than $\minPointFrustum = 10$ points, which can happen for very distant objects.

Each Frustum Proposal is then processed by the Frustum Localizer (Frustum PointNet \cite{frustum-pointnet}), a 3D Localization model that predicts a 3D bounding box from a single frustum. We then assign the semantic label of the 2D detection to the recovered 3D bounding box. We also shrink the confidence score by the IoU with the 2D detection to penalize inaccurate bounding boxes:
\begin{equation} \label{eq:iou-downweight}
    s_j^{3d} = s_{j}^{2d} \cdot IoU(\bboxGeneric{proj}_j, \bboxImg{j}),
\end{equation}
where $s_{j}^{2d}$ is the confidence score of the RGB detection, and $\bboxGeneric{proj}_j$ is the projection of the localized object by the Frustum Localizer in the image plane. We confirm the new detection if and only if the IoU between its projection and the corresponding 2D bounding box is higher than a threshold $\iouThrDetRec$.

We can also use instance segmentation in the RGB branch to generate frustum proposals that include only the 3D points projected inside the segmentation mask (Figure \ref{fig:frustums:mask_frustum}), which typically lie within the object of interest. In practice, we extract the frustum from the 2D bounding box and add an additional binary channel encoding the mask information for each point (Figure \ref{fig:frustums:mask_frustum_one_hot}). This strategy preserves relevant points in case of inconsistent masks and enriches the frustum representation with finer semantic cues.


\subsubsection{Semantic Fusion} \label{sec:sem-fusion}

\begin{figure}[t]
    \centering
    \includegraphics[width=0.3\linewidth]{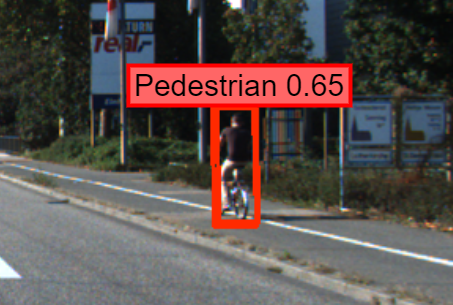}\label{fig:semantic:lidar}
    \includegraphics[width=0.3\linewidth]{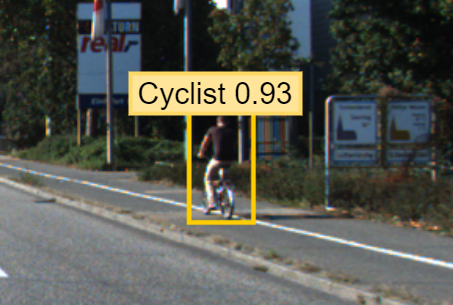}\label{fig:semantic:cam}
    \caption{\textbf{Left}: a Cyclist wrongly classified as Pedestrian by the LiDAR branch. \textbf{Right}: the Cyclist is correctly classified by the RGB branch.}
    \label{fig:semantic}
\end{figure}

The Semantic Fusion module, described in Algorithm \ref{alg:semantic_fusion}, combines both the labels to enforce semantic consistency and the confidence scores of LiDAR and RGB detections. In fact, it may happen that 2D and 3D bounding boxes matched by the Bounding Box Matching module are associated to different semantic labels in $\clsSet$ (Figure \ref{fig:semantic}). The Semantic Fusion module takes as input the set $\MatchingFinal = \Matching \cup \MatchingRecover$, including both matched ($\Matching$) and recovered ($\MatchingRecover$) detections. Since RGB images better capture semantic details, the RGB detector is expected to be better at defining object classes. Thus, we associate the labels estimated from the RGB branch to each matched LiDAR bounding box (Line 5), as in \cite{long_tailed_late_fusion}. When the two predicted classes are different, we assign the label and the confidence of the RGB detector to the 3D detection (Line 9). Instead, when the predicted classes are equal, we follow the probabilistic ensemble framework in \cite{chen2022multimodal}, assuming conditional independence between modalities, and define the final detection confidence score $s^{\prime}$ for class $\lambda \in \clsSet$: $s^{\prime}(\lambda) \propto s^{3d} \cdot s^{2d} \mathbin{/} p(\lambda)$, where $p(\lambda)$ is the class prior, treated as a uniform in this work.

\begin{compactalgorithm}
\caption{Semantic Fusion} \label{alg:semantic_fusion}
\hspace*{\algorithmicindent} \textbf{Input}: Matching detections $\MatchingFinal$\\
\hspace*{\algorithmicindent} \textbf{Output}: Final detection output $\bboxSet$
\begin{algorithmic}[1]
\Function{SemanticFusion}{$\MatchingFinal$}
\State $\bboxSet \gets \emptyset$
\For{\texttt{$(\bboxLidar{j}, s_j^{3d}, \cls{j}^{3d}, \bboxImg{j}, s_j^{2d}, \cls{j}^{2d}) \in \MatchingFinal$}}
    \State $\cls{j}^{\prime} \gets \cls{j}^{2d}$
    \If{$\cls{j}^{2d} = \cls{j}^{3d}$}
        \State $s_j^{\prime} \gets \Call{ProbabilisticEnsemble}{s_j^{3d}, \cls{j}^{3d}, s^{2d}, \cls{j}^{\prime}}$
    \Else
        \State $s_j^{\prime} \gets s_j^{2d}$
    \EndIf
    \State $\bboxSet \gets \bboxSet \cup (\bboxLidar{j}, s_j^{\prime}, \cls{j}^{\prime})$
\EndFor
\State \Return $\bboxSet$
\EndFunction
\end{algorithmic}
\end{compactalgorithm}

\subsection{The Stereo View case} \label{sec:stereo-view-case}

We now revisit the single-view approach for the case in which the input consists of stereo images. In short, from a stereo pair we can leverage two sets of detections, one for the left view and one from the right view. Thus, we consider a 3D detection from the LiDAR branch as True Positive (TP) when it matches a 2D bounding box at least in one image. In the Detection Recovery, we leverage the epipolar geometry of the stereo pair to pair unmatched bounding boxes from the two views and intersect the frustums from the two matched bounding boxes to further reduce the 3D search space.

\subsubsection{Stereo RGB branch}

The input of the stereo RGB branch is composed by stereo images ($\stereoPair{})$. Thus, the output is the pair of sets $(\bboxSetImgStereo{l}, \bboxSetImgStereo{r})$, where $\bboxSetImgStereo{l}$ and $\bboxSetImgStereo{r}$ are the set of bounding boxes in the left and right images, respectively, defined as in the single-view case.

\subsubsection{Stereo Bounding Box Matching}

In the stereo view setting, the Bounding Box Matching module can leverage 2D bounding boxes from both left and right views. In particular, we repeat the previously described method for both the images $\img{}^l$ and $\img{}^r$ and we confirm the LiDAR detections when these are matched with 2D bounding boxes in any of the two images. Table \ref{tab:matching_summary} summarizes the process for one pair of stereo images.

\begin{table}[]
    \centering
    \caption{Bounding box matching summary for $\numImg = 1$ stereo views}
    \resizebox{0.7\textwidth}{!}{%
    \begin{tabular}{|>{\centering\arraybackslash}p{2cm}|>{\centering\arraybackslash}p{2cm}|>{\centering\arraybackslash}p{2cm}||>{\centering\arraybackslash}p{5cm}|}
    \hline
    LiDAR & RGB left & RGB right & Comment \\
    \hline
    \xmark & \xmark & \xmark & No detection \\
    \hline
    \xmark & \xmark & \cmark & RGB right False Positive \\
    \hline
    \xmark & \cmark & \xmark & RGB left False Positive \\
    \hline
    \xmark & \cmark & \cmark & Detection Recovery \\
    \hline
    \cmark & \xmark & \xmark & LiDAR False Positive \\
    \hline
    \cmark & \xmark & \cmark & Semantic Fusion \\
    \hline
    \cmark & \cmark & \xmark & Semantic Fusion \\
    \hline
    \cmark & \cmark & \cmark & Semantic Fusion \\
    \hline
    \end{tabular}
    \label{tab:matching_summary}
    }
\end{table}

\subsubsection{Stereo Detection Recovery} \label{sec:recovery-stereo}

RGB detections from both left and right image that are not matched with any 3D detection are fed to the stereo Detection Recovery module. At a high level, the stereo Detection Recovery module performs three steps. First, we match each bounding box $\bboxImg{l} \in \bboxSetImgUnmatchStereo{l}$ in the left image with possibly one bounding box $\bboxImg{r} \in \bboxSetImgUnmatchStereo{r}$ in the right view, exploiting two-view geometry constraints. Then, we extract Frustum Proposals from each pair of matched detections and execute the Frustrum Localizer on their intersection.

\begin{figure}[t]
    \centering
    \begin{minipage}{0.45\textwidth}
        \centering
        \subfloat[2D detections on the left image\label{fig:det_recovery:left_bboxes}]{\includegraphics[width=\linewidth]{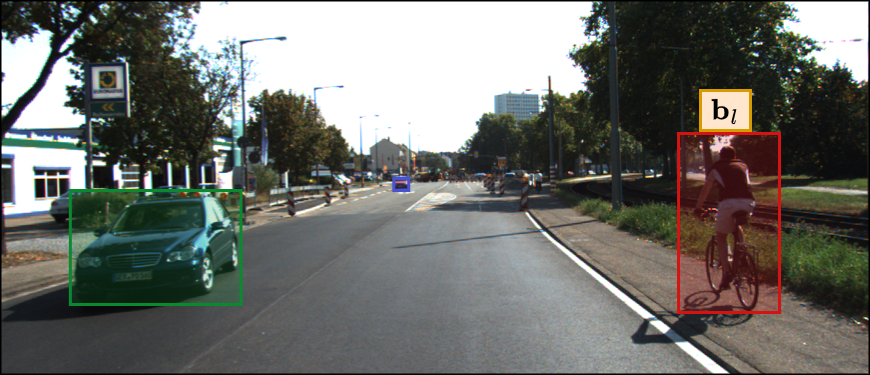}}
        \vfill
        \subfloat[2D detections and epipolar lines on the right image\label{fig:det_recovery:right_epipolar}]{\includegraphics[width=\linewidth]{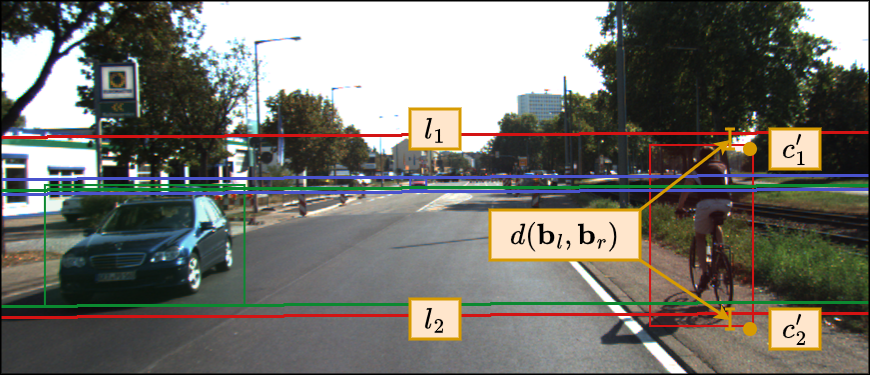}}
    \end{minipage}
    \hfill
    \begin{minipage}{0.45\textwidth}
         \subfloat[Frustum Point Cloud obtained by selecting the points inside the pair of frustums given by the two assigned detections.\label{fig:det_recovery:frustum}]{\includegraphics[width=\linewidth]{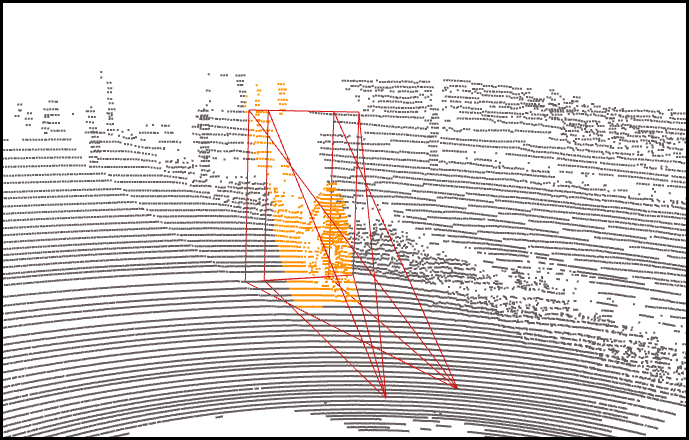}}
    \end{minipage}
    \caption{Illustration of the Frustum Proposals, obtained from the Detection Recovery module assignment procedure between two detections $(\bboxImg{l}, \bboxImg{r})$ belonging to stereo images $\img{}^l$ and $\img{}^r$.}
    \label{fig:det_recovery}
\end{figure}

To match left and right bounding boxes, we design an epipolar assignment procedure. Given a bounding box $\bboxImg{l}$ detected on the left image, we compute the epipolar lines $(l_1, l_2)$ corresponding to its top left and bottom right corners $(c_1, c_2)$ on the right image as $l_k = \fMat{l}{r} c_k, \quad k \in \{1, 2\}$, where $\fMat{l}{r}$ is the fundamental matrix between the two images, computed as:
\begin{equation} \label{eq:fundamental_matrix}
    \fMat{l}{r} = K_l^{-T} E K_r^{-1} = K_l^{-T} [t]_{\times} R K_r^{-1},
\end{equation}
where $K_l$ and $K_r$ are the intrinsic matrices of the two cameras, $R$ and $t$ are the relative rotation and translation between the cameras, respectively.
When the stereo pair is rectified as in Figure \ref{fig:det_recovery:right_epipolar}, the epipolar lines are horizontal. Ideally, the same corners $(c_1^\prime, c_2^\prime)$ of a bounding box $\bboxImg{r}$ corresponding to the same object in the right image should belong to these two epipolar lines. However, the predictions of the RGB branch may have small inconsistencies, as shown in Figures \ref{fig:det_recovery:left_bboxes} and \ref{fig:det_recovery:right_epipolar}, but still the corners $(c_1^\prime, c_2^\prime)$ are expected to be close to the epipolar lines $(l_1, l_2)$ defined by the bounding box in the other image. This is illustrated for the cyclist in Figure \ref{fig:det_recovery:right_epipolar}.

This motivates our cost function $d(\cdot,\cdot)$ for matching $\bboxImg{l}$ and $\bboxImg{r}$ as the sum of the Euclidean distances $\tilde{d}(\cdot,\cdot)$ between each corner of $\bboxImg{r}$ and the epipolar lines of the corresponding corner of $\bboxImg{l}$:
\begin{equation} \label{eq:distance_epipolar}
    d(\bboxImg{l}, \bboxImg{r}) = \tilde{d}(l_1, c_1^\prime) + \tilde{d}(l_2, c_2^\prime).
\end{equation}
The assignment problem is similar to \eqref{eq:linear_sum_assign}, but here we minimize this distance instead of maximizing the IoU. As before, we solve the assignment problem using the Jonker-Volgenant algorithm, to get matches $\mathcal{M}_{2d}$ as output. 

Given the matched pairs of 2D detections in $\mathcal{M}_{2d}$ from the stereo views, we extract 3D Frustum Proposals by back-projecting each detection in a frustum and intersecting the two frustums from both views (Figure \ref{fig:det_recovery:frustum}). The same Frustum Localizer used in the single-view setting produces a 3D bounding box from the Frustum Proposal. We then assign to the 3D bounding box the estimated label and the score of the most confident RGB detection and down-weight the confidence score by the IoU with the 2D detections as:
\begin{equation} \label{eq:iou-downweight-stereo}
    s^{3d} = s_{RGB} \cdot IoU(\bboxGeneric{proj}_l, \bboxImg{l}) \cdot IoU(\bboxGeneric{proj}_r, \bboxImg{r}),
\end{equation}
where $s_{RGB}$ is the confidence score of the most confident RGB detection, and $(\bboxGeneric{proj}_l, \bboxGeneric{proj}_r)$ are the projected 2D bounding boxes in the two image planes of the bounding box predicted by the Frustum Localizer. We then discard the recovered 3D detections that do not satisfy the following condition:
\begin{equation}
    \min(IoU(\bboxGeneric{proj}_l, \bboxImg{l}) \cdot IoU(\bboxGeneric{proj}_r, \bboxImg{r})) \geq \iouThrDetRec.
\end{equation}

\subsubsection{Stereo Semantic Fusion}
The Semantic Fusion module for the stereo-view setting is similar to the single-view one, with the difference that each LiDAR 3D detection can be associated with the labels from one or two 2D RGB bounding boxes when it is matched in both stereo images. When the number of matched RGB detections is one, this reduces to the same procedure as the single-view setting. Differently, when 2D detections from both images are matched, we take the label of the most confident-matched RGB detection to assign the label to the matched 3D bounding box, and fuse the scores as in Section \ref{sec:sem-fusion}, but with an additional term for the second image: $s^{\prime}(\lambda) \propto s^{3d} \cdot s_l^{2d} \cdot s_r^{2d} \mathbin{/} p(\lambda)$.

\section{Experiments}
\label{sec:exp}
We test LCF3D with KITTI \cite{kitti_dataset} and nuScenes \cite{nuscenes2019} datasets, comparing our solution with state-of-the-art LiDAR-based and multimodal methods for 3D Object Detection. KITTI and nuScenes provide very different settings for both the LiDAR sensor and the type of RGB cameras employed, resulting in domain shifts when one dataset is used for training and the other for inference. We test the Domain Generalization performance of LCF3D in these settings.

\subsection{Datasets} \label{sec:experimental_setup}
The \textbf{KITTI} object detection dataset \cite{kitti_dataset} includes data from a 64-beam LiDAR and a stereo RGB camera pair (in the single-view setup, only the left camera is used). Following the protocol in \cite{mv3d}, we split the training set into 3712 training and 3769 validation samples, and adopt the official evaluation scheme with three difficulty levels: easy (fully visible, nearby objects), moderate (partially occluded or more distant), and hard (small or heavily occluded objects).
 
The \textbf{nuScenes} dataset \cite{nuscenes2019} provides a comprehensive sensor suite with one 32-beam LiDAR and six non-overlapping RGB cameras covering the full field of view. To train the RGB branch, we additionally use nuImages, a complementary dataset of 93k images sharing the same sensor setup and including instance-segmentation labels. We train 2D detection models on both nuScenes and nuImages, and instance segmentation models on nuImages only.
\subsection{Figure of merits}

We consider both  KITTI and nuScenes metrics. For KITTI, we use the 3D Average Precision (AP) and we consider the Car, Pedestrian and Cyclist classes. For nuScenes,  we use the per-class Average Precision (AP), the mean Average Precision (mAP) and the nuScenes Detection Score (NDS) using all the 10 classes. More details are in \cite{nuscenes2019}. We measure our inference speed on an A100 GPU.

To assess Domain Generalization, we follow the approach  by DG-BEV \cite{wang2023towards} to compute a variant of the original NDS. Indeed, the original NDS aggregates six metrics, including mAP, mATE, mASE, mAOE, mAVE and mAAE. As velocity and attributes are present only in nuScenes, we adopt the figure of merit $NDS^{\hat{*}}$, proposed by DG-BEV \cite{wang2023towards} to not involve mAVE and mAAE. $NDS^{\hat{*}}$ is computed as:
\begin{equation}\label{eq:nds_hat}
    NDS^{\hat{*}} = \frac{1}{6} (3 mAP + \sum_{mTP \in \mathbb{T}\mathbb{P}} (1 - min(1, mTP) )
\end{equation}
where $\mathbb{T}\mathbb{P} = \{mATE, mASE, mAOE\}$. While nuScenes provides annotations in the ring-view, KITTI is limited to the front-view. Thus, for a fair comparison, in these experiments, we limit the evaluation of nuScenes models only to the front-view, i.e., the field of view of the front camera.

\subsection{Competitors}

We compare LCF3D against state-of-the-art multimodal models \cite{frustum-pointnet,li2023logonet,clocs_late_fusion,virconv,zhang2022catdetcontrastivelyaugmentedtransformer,lin2023mlf} on the KITTI validation set. Results and inference speeds are taken from the corresponding official papers. 

To test Domain Generalization, we compare LCF3D, configured with PointPillars \cite{pointpillars} in the LiDAR branch and FasterRCNN \cite{faster_rcnn} in the RGB branch, against a representative early fusion technique (MVXNet \cite{mvx-net}) and an intermediate fusion solution (BEVFusion \cite{liu2023bevfusion}). We do not include Domain Adaptation methods in the analysis, as they pursue a different objective, whereas our goal is to evaluate generalization to unseen domains without any adaptation.

\subsection{Implementation Details} \label{sec:implementation-details}

\subsubsection{LCF3D configuration on KITTI} \label{sec:kitti-setup}

\textbf{LiDAR Detectors}. We test the pre-trained PointPillars \cite{pointpillars}, PV-RCNN \cite{pv-rcnn} and PartA2 \cite{partA2} models by MMDetection3D and train SECOND \cite{second}  on the KITTI training split for 80 epochs using standard augmentations (object noise, BEV random flip, and ground-truth sampling).

\textbf{RGB Detectors}. A Faster R-CNN \cite{faster_rcnn} with ResNet101-FPN backbone taken from MMDetection is fine-tuned on the left KITTI images.

\textbf{Frustum Localizer}. 
Frustum PointNet \cite{frustum-pointnet} is re-implemented within MMDetection3D and trained on frustums from 2D ground-truth boxes, with separate models for single- and stereo-view setups.

\subsubsection{LCF3D configuration on nuScenes} \label{sec:nuscenes-setting}

\textbf{LiDAR Detectors}. We use pre-trained models (PointPillars \cite{pointpillars}, SSN \cite{zhu2020ssn} and CenterPoint \cite{yin2021center}) on the nuScenes dataset from the MMDetection3D framework.

\textbf{RGB Detectors}. For Object Detection, we train a Faster RCNN \cite{faster_rcnn} model with ResNet50 as backbone and an FPN as the neck, and a DDQ-DETR \cite{zhang2023dense} network with Swin-L as backbone. For Instance Segmentation, we use DetectorRS \cite{qiao2021detectors}. All the models were trained using MMDetection's framework for 12 epochs.

\textbf{Frustum Localizer}. We train a Frustum Pointnet of single-view Frustum Proposals extracted from 2D bounding boxes obtained through the projection of the 3D ones in the images, as in \ref{sec:kitti-setup}. We also train the Frustum Localizer on frustums extracted from 2D instance segmentation masks. We use the trained DetectorRS model to generate instance masks that we match with 3D ground truths using Bounding Box Matching.

\subsection{Results on KITTI}

\begin{table}[t]
\centering
\caption{Comparison with single modal detectors (3D AP) on the KITTI val set. Blue denotes the \textcolor{NavyBlue}{best overall performance}, while green denotes the \textcolor{ForestGreen}{best performance among the rows} with the same 3D detector on the LiDAR branch. Rows having an empty RGB branch and RGB setting denote the single-modal LiDAR solution.}
\resizebox{0.9\textwidth}{!}{%
\begin{tabular}{|c|c|c|>{\centering\arraybackslash}p{1.2cm}|>{\centering\arraybackslash}p{1.2cm}|>{\centering\arraybackslash}p{1.2cm}|>{\centering\arraybackslash}p{1.2cm}|>{\centering\arraybackslash}p{1.2cm}|>{\centering\arraybackslash}p{1.2cm}|>{\centering\arraybackslash}p{1.2cm}|>{\centering\arraybackslash}p{1.2cm}|>{\centering\arraybackslash}p{1.2cm}|}
\hline
\multirow{2}{*}{LiDAR branch} & \multirow{2}{*}{RGB branch} & \multirow{2}{*}{RGB setting} & \multicolumn{3}{c|}{Car $AP_{3d} \uparrow$ } & \multicolumn{3}{c|}{Pedestrian $AP_{3d} \uparrow$} & \multicolumn{3}{c|}{Cyclist $AP_{3d} \uparrow$} \\
\cline{4-12}
& & & Easy & Mod. & Hard & Easy & Mod. & Hard & Easy & Mod. & Hard \\
\hhline{|=|=|=|=|=|=|=|=|=|=|=|=|}
SECOND & - & - & 87.83 & 78.46 & 73.75 & 59.12 & 52.78 & 47.41 & 75.58 & 61.73 & 58.18 \\
\hline
SECOND & Faster RCNN & Stereo & \textbf{\textcolor{ForestGreen}{88.41}} & \textbf{\textcolor{ForestGreen}{79.45}} & \textbf{\textcolor{ForestGreen}{74.35}} & 65.98 & 59.73 & 53.47 & 85.24 & 73.44 & 69.23 \\
\hline
SECOND & Faster RCNN & Single & 88.31 & 79.04 & 74.29 & \textbf{\textcolor{ForestGreen}{66.33}} & \textbf{\textcolor{ForestGreen}{61.84}} & \textbf{\textcolor{ForestGreen}{55.80}} & \textbf{\textcolor{ForestGreen}{85.49}} & \textbf{\textcolor{ForestGreen}{74.24}} & \textbf{\textcolor{ForestGreen}{69.54}} \\
\hhline{|=|=|=|=|=|=|=|=|=|=|=|=|}
PointPillars & - & - & 88.52 & 79.29 & 76.34 & 57.27 & 51.00 & 46.44 & 83.88 & 62.77 & 59.50 \\
\hline
PointPillars & Faster RCNN & Stereo & \textbf{\textcolor{ForestGreen}{89.45}} & \textbf{\textcolor{ForestGreen}{80.29}} & 77.24 & \textbf{\textcolor{ForestGreen}{70.38}} & 63.98 & 58.76  & \textbf{\textcolor{ForestGreen}{88.07}} & 73.88 & 69.07 \\
\hline
PointPillars & Faster RCNN & Single & 89.17 & 80.21 & \textbf{\textcolor{ForestGreen}{77.44}} & 69.81 & \textbf{\textcolor{ForestGreen}{64.66}} & \textbf{\textcolor{ForestGreen}{59.90}} & 87.13 & \textbf{\textcolor{ForestGreen}{75.74}} & \textbf{\textcolor{ForestGreen}{70.86}} \\
\hhline{|=|=|=|=|=|=|=|=|=|=|=|=|}
PartA2 & - & - & 92.45 & 82.88 & 80.64 & 60.61 & 53.59 & 48.86 & 90.45 & 70.17 & 65.52 \\
\hline
PartA2 & Faster RCNN & Stereo & \textbf{\textcolor{NavyBlue}{92.98}} & \textbf{\textcolor{ForestGreen}{83.80}} & \textbf{\textcolor{ForestGreen}{81.37}} & 72.44 & 65.52 & 58.98 & 94.01 & 79.39 & 74.28 \\
\hline
PartA2 & Faster RCNN & Single & 92.75 & 83.44 & 81.16 & \textbf{\textcolor{ForestGreen}{72.67}} & \textbf{\textcolor{ForestGreen}{67.35}} & \textbf{\textcolor{ForestGreen}{61.11}} & 93.64 & \textbf{\textcolor{NavyBlue}{79.94}} & \textbf{\textcolor{NavyBlue}{74.99}} \\
\hhline{|=|=|=|=|=|=|=|=|=|=|=|=|}
PV-RCNN & - & - & 91.82 & 84.53 & 82.42 & 66.72 & 59.27 & 54.31 & 90.36 & 73.26 & 69.36 \\
\hline
PV-RCNN & Faster RCNN & Stereo & \textbf{\textcolor{ForestGreen}{92.93}} & \textbf{\textcolor{NavyBlue}{86.31}} & 83.34 & 73.86 & 68.44 & 63.61 & 91.01 & 77.25 & 72.01 \\
\hline
PV-RCNN & Faster RCNN & Single & 92.81 & 86.30 & \textbf{\textcolor{NavyBlue}{83.63}} & \textbf{\textcolor{NavyBlue}{73.89}} & \textbf{\textcolor{NavyBlue}{68.60}} & \textbf{\textcolor{NavyBlue}{63.80}} & \textbf{\textcolor{NavyBlue}{91.88}} & \textbf{\textcolor{ForestGreen}{77.47}} & \textbf{\textcolor{ForestGreen}{72.44}} \\
\hline
\end{tabular}
}
\label{table:single_mod_comp_3d}
\end{table}

Table \ref{table:single_mod_comp_3d} compares our performance with LiDAR-based 3D Object Detectors. We group the rows based on the specific 3D Detector used in the LiDAR branch and report the metrics of both single and stereo-view settings. LCF3D significantly outperforms single-modal detectors on highly imbalanced classes (Pedestrians and Cyclists) in all difficulty levels, especially when these are distant from the sensor (moderate and hard). In the easy case, stereo vision is more reliable, while single-view yields better results in the moderate and hard cases for both Pedestrians and Cyclists. 
Indeed, the stereo setup improves detection of nearby objects but is less effective for distant ones. The epipolar-based matching requires consistent 2D detections in both views, which benefits easy cases but limits Detection Recovery when objects are far or partially occluded, resulting in fewer and sparser frustum proposals than in the single-view setting.

In Table \ref{table:multi_modal_comparison}, we collate the results of LCF3D with other multi-modal solutions on the KITTI validation set. VirConv \cite{virconv} still outperforms our method for Cars, but for Pedestrians and Cyclists we achieve state-of-the-art results by combining PV-RCNN on the LiDAR branch and Faster R-CNN on the RGB branch in the single-view setting (\emph{LCF3D-Single (FR + PV)} in Table \ref{table:multi_modal_comparison}). Moreover, using PointPillars, our approach ensure significantly lower computational times compared to other multi-modal solutions, while remaining competitive. Thus, although the comparison may not be based on the same hardware architectures, it remains indicative, as we found that our measured inference speeds of the PointPillars and PV-RCNN are in line with the official ones.

\begin{table}[t]
\centering
\caption{Performance comparison with multi-modal solutions on the KITTI val set, using Faster RCNN (FR) in the RGB branch and PointPillars (PP) or PV-RCNN (PV) in the LiDAR branch. Best results are in \textbf{bold}, second best results are \underline{underlined}. Inference speed is taken from the original publications, when available.}
\resizebox{0.9\textwidth}{!}{%
\begin{tabular}{|c|>{\centering\arraybackslash}p{1.2cm}|>{\centering\arraybackslash}p{1.2cm}|>{\centering\arraybackslash}p{1.2cm}|>{\centering\arraybackslash}p{1.2cm}|>{\centering\arraybackslash}p{1.2cm}|>{\centering\arraybackslash}p{1.2cm}|>{\centering\arraybackslash}p{1.2cm}|>{\centering\arraybackslash}p{1.2cm}|>{\centering\arraybackslash}p{1.2cm}|>{\centering\arraybackslash}p{1.2cm}|}
\hline
\multirow{2}{*}{Detector} & \multirow{2}{*}{\makecell{Speed \\ (FPS$^*$)}} & \multicolumn{3}{c|}{Car $AP_{3d} \uparrow$} & \multicolumn{3}{c|}{Pedestrian $AP_{3d} \uparrow$} & \multicolumn{3}{c|}{Cyclist $AP_{3d} \uparrow$} \\
\cline{3-11}
& & Easy & Mod. & Hard & Easy & Mod. & Hard & Easy & Mod. & Hard \\
\hline
CLOCs-PVCas \cite{clocs_late_fusion} & - & 89.49 & 79.31 & 77.36 & 62.88 & 56.20 & 50.10 & 87.57 & 67.92 & 63.67 \\
\hline
Frustum PointNet \cite{frustum-pointnet} & 5.9 & 83.76 & 70.92 & 63.65 & 70.00 & 61.32 & 53.59 & 77.15 & 56.49 & 53.37 \\
\hline
Frustum PointPillars \cite{frustum_pointpillars} & 14.3 & 88.90 & 79.28 & 78.07 & 66.11 & 61.89 & 56.91 & 87.54 & 72.78 & 66.07 \\
\hline
PointPainting \cite{point-painting} & - & 88.38 & 77.74 & 76.76 & 69.38 & 61.67 & 54.58 & 85.21 & 71.62 & 66.98 \\
\hline
MVXNet \cite{mvx-net} & - & 88.48 & 78.75 & 74.34 & 58.27 & 55.51 & 51.83 & 79.15 & 63.25 & 60.56 \\
\hline
CAT-Det \cite{zhang2022catdetcontrastivelyaugmentedtransformer} & 10.2 & 90.12 & 81.46 & 79.15 & \textbf{74.08} & 66.35 & 58.92 & 87.64 & 72.82 & 68.20 \\
\hline
VirConv-T \cite{virconv} & 10.2 & \textbf{94.98} & \textbf{89.96} & \textbf{88.13} & 73.32 & 66.93 & 60.38 & 90.04 & 73.90 & 69.06 \\
\hline
LoGoNet & - & 92.04 & 85.04 & \underline{84.31} & 70.20 & 63.72 & 59.46 & \textbf{91.74} & 75.35 & \textbf{72.42} \\
\hline
MLF-DET-V & 10.8 & 89.70 & \underline{87.31} & 79.34 & 71.15 & \textbf{68.50} & 61.72 & 86.05 & 72.14 & 65.42 \\
\hhline{|=|=|=|=|=|=|=|=|=|=|=|}
LCF3D-Single (FR + PP) & 30.4 & 89.30 & 80.03 & 77.23 & 69.81 & 64.66 & 59.90 & 87.13 & 75.74 & 70.86 \\
\hline
LCF3D-Single (FR + PV) & 10.5 & 92.44 & 85.99 & 83.54 & 73.43 & \underline{68.18} & \textbf{63.56} & 89.83 & \textbf{77.27} & \underline{72.17} \\
\hline
LCF3D-Stereo (FR + PV) & 10.1 & \underline{92.95} & 86.09 & 83.32 & \underline{73.87} & 67.40 & 62.67 & \underline{91.01} & 77.25 & 72.01 \\
\hline
\end{tabular}
}
\label{table:multi_modal_comparison}
\end{table}

\subsection{Results on nuScenes}
Results on the validation set of nuScenes are collected in Table \ref{table:nuScenes-results}. The rundown of the experiment is similar to the single-view case of KITTI. The benefits of LCF3D are marginal for Cars. However, the improvements are very noticeable for imbalanced classes such as Bicycles and Motorcycles, as well as for classes associated with small objects like Traffic Cones and Barriers. The advantages are less evident for Pedestrians, as they do not constitute an imbalanced class, as confirmed by the  strong performance of the LiDAR branch alone. However, for this category, our method still provides significant improvements over PointPillars and SSN. Interestingly, we do not observe improvements for CenterPoint on Pedestrians, suggesting that the RGB detectors we employed do not outperform the baseline of CenterPoint in this class.
The modular design of LCF3D makes it compatible with different RGB and LiDAR detectors without architectural modification. We verified this by combining various backbones (e.g., Faster R-CNN, DDQ-DETR, DetectorRS PointPillars, SSN, CenterPoint), and observed consistent improvements across all setups, as shown in Table \ref{table:nuScenes-results}. Overall, these results confirm the generalization ability of LCF3D while showing that its performance naturally depends on the reliability of the underlying single-modal detectors.

\begin{table}[t]
\centering
\caption{Results on the nuScenes validation set. Con.V., Pedes. Motor. and TC are abbreviations for Construction Vehicles, Pedestrians, Motorcycles and Traffic Cones, respectively. Please note that DetectorRS (*) is trained only on nuImages, while Faster RCNN and DDQ are also trained on nuScenes. In green we report the best results among variants with the same LiDAR detector, in blue the best overall performance.}
\resizebox{\textwidth}{!}{%
\begin{tabular}{|c|c||>{\centering\arraybackslash}p{1.4cm}|>{\centering\arraybackslash}p{1.4cm}||>{\centering\arraybackslash}p{1.4cm}|>{\centering\arraybackslash}p{1.4cm}|>{\centering\arraybackslash}p{1.4cm}|>{\centering\arraybackslash}p{1.4cm}|>{\centering\arraybackslash}p{1.4cm}|>{\centering\arraybackslash}p{1.4cm}|>{\centering\arraybackslash}p{1.4cm}|>{\centering\arraybackslash}p{1.4cm}|>{\centering\arraybackslash}p{1.4cm}|>{\centering\arraybackslash}p{1.4cm}|}
\hline
LiDAR branch & RGB branch & mAP $\uparrow$ & NDS $\uparrow$ & Car $\uparrow$ & Truck $\uparrow$ & Bus $\uparrow$ & Trailer $\uparrow$ & Con.V. $\uparrow$ & Pedes. $\uparrow$ & Motor. $\uparrow$ & Bicycle $\uparrow$ & TC $\uparrow$ & Barrier $\uparrow$ \\
\hhline{|=|=|=|=|=|=|=|=|=|=|=|=|=|=|}
PointPillars \cite{pointpillars} & - & 0.390 & 0.526 & 0.797 & 0.354 & 0.427 & 0.256 & 0.050 & 0.682 & 0.382 & 0.105 & 0.334 & 0.515 \\
\hline
PointPillars & FasterRCNN & 0.533 & 0.588 & 0.797 & 0.430 & 0.474 & 0.235 & 0.157 & 0.812 & 0.612 & 0.512 & 0.697 & 0.606 \\
\hline
PointPillars & DDQ & \textbf{\textcolor{ForestGreen}{0.570}} & \textbf{\textcolor{ForestGreen}{0.609}} & \textbf{\textcolor{ForestGreen}{0.813}} & \textbf{\textcolor{ForestGreen}{0.490}} & \textbf{\textcolor{ForestGreen}{0.551}} & \textbf{\textcolor{ForestGreen}{0.307}} & \textbf{\textcolor{ForestGreen}{0.198}} & \textbf{\textcolor{ForestGreen}{0.830}} & \textbf{\textcolor{ForestGreen}{0.655}} & \textbf{\textcolor{ForestGreen}{0.536}} & \textbf{\textcolor{ForestGreen}{0.701}} & 0.618 \\
\hline
PointPillars & DetectorRS* & 0.533 & 0.585 & 0.796 & 0.454 & 0.454 & 0.210 & 0.153 & 0.779 & 0.615 & 0.518 & 0.626 & \textbf{\textcolor{ForestGreen}{0.653}} \\
\hhline{|=|=|=|=|=|=|=|=|=|=|=|=|=|=|}
SSN \cite{zhu2020ssn} & - & 0.459 & 0.577 & 0.827 & 0.518 & 0.611 & 0.314 & 0.158 & 0.666 & 0.473 & 0.219 & 0.271 & 0.536 \\
\hline
SSN & FasterRCNN & 0.570 & 0.627 & 0.821 & 0.532 & 0.606 & 0.243 & 0.215 & 0.813 & 0.637 & 0.552 & 0.664 & 0.621 \\
\hline
SSN & DDQ & \textbf{\textcolor{ForestGreen}{0.607}} & \textbf{\textcolor{ForestGreen}{0.648}} & \textbf{\textcolor{ForestGreen}{0.837}} & \textbf{\textcolor{ForestGreen}{0.591}} & \textbf{\textcolor{ForestGreen}{0.668}} & \textbf{\textcolor{ForestGreen}{0.311}} & \textbf{\textcolor{ForestGreen}{0.251}} & \textbf{\textcolor{ForestGreen}{0.833}} & \textbf{\textcolor{ForestGreen}{0.682}} & \textbf{\textcolor{ForestGreen}{0.575}} & \textbf{\textcolor{ForestGreen}{0.676}} & 0.642 \\
\hline
SSN & DetectorRS* & 0.560 & 0.614 & 0.822 & 0.536 & 0.625 & 0.222 & 0.198 & 0.779 & 0.629 & 0.549 & 0.576 & \textbf{\textcolor{ForestGreen}{0.660}} \\
\hhline{|=|=|=|=|=|=|=|=|=|=|=|=|=|=|}
CenterPoint \cite{yin2021center} & - & 0.554 & 0.641 & 0.845 & 0.523 & 0.666 & 0.359 & 0.156 & 0.827 & 0.529 & 0.344 & 0.638 & 0.653 \\
\hline
CenterPoint & FasterRCNN & 0.609 & 0.661 & 0.829 & 0.542 & 0.620 & 0.335 & 0.233 & 0.847 & 0.654 & 0.591 & \textbf{\textcolor{NavyBlue}{0.747}} & 0.689 \\
\hline
CenterPoint & DDQ & \textbf{\textcolor{NavyBlue}{0.635}} & \textbf{\textcolor{NavyBlue}{0.674}} & \textbf{\textcolor{NavyBlue}{0.846}} & \textbf{\textcolor{NavyBlue}{0.592}} & \textbf{\textcolor{NavyBlue}{0.724}} & \textbf{\textcolor{NavyBlue}{0.382}} & \textbf{\textcolor{NavyBlue}{0.251}} & \textbf{\textcolor{NavyBlue}{0.865}} & \textbf{\textcolor{NavyBlue}{0.680}} & \textbf{\textcolor{NavyBlue}{0.594}} & 0.739 & 0.675 \\
\hline
CenterPoint & DetectorRS* & 0.609 & 0.658 & 0.832 & 0.550 & 0.663 & 0.302 & 0.215 & 0.820 & 0.655 & 0.580 & 0.730 & \textbf{\textcolor{NavyBlue}{0.740}} \\
\hline
\end{tabular}
}
\label{table:nuScenes-results}
\end{table}

\subsection{Ablation Studies}

We assess each LCF3D module, using its modularity to form selective baselines.
Using Bounding Box Matching alone produces matched 3D detections with classes predicted by the LiDAR branch. Adding Semantic Fusion replaces the class labels with those from the RGB branch and adjusts the detection scores. Semantic Fusion depends on matched detections from both modalities, thus cannot be tested alone.
The Detection Recovery only baseline represents a pure cascade-fusion setup, where 3D boxes are generated solely from frustum proposals built on all the RGB detections, as there is no LiDAR branch. This modular evaluation allows direct quantification of each component’s contribution.

\subsubsection{Modules Contribution}

\begin{table}[t]
\centering
\caption{Ablation studies on the KITTI val set (Single View).}
\resizebox{0.55\textwidth}{!}{%
\begin{tabular}{|>{\centering\arraybackslash}p{2cm}|>{\centering\arraybackslash}p{2cm}|>{\centering\arraybackslash}p{2cm}|>{\centering\arraybackslash}p{1.1cm}|>{\centering\arraybackslash}p{1.1cm}|>{\centering\arraybackslash}p{1.1cm}|>{\centering\arraybackslash}p{1.1cm}|}
\hline
\multirow{2}{*}{Bbox Match.} & \multirow{2}{*}{Det. Recovery} & \multirow{2}{*}{Sem. Fusion} & \multicolumn{3}{c|}{Overall $AP_{3d} \uparrow$} & \multirow{2}{*}{\makecell{Speed \\ (FPS) $\uparrow$}}\\
\cline{4-7}
& & & Easy & Mod. & Hard  &\\
\hhline{|=|=|=|=|=|=|=|}
\xmark & \xmark & \xmark & 76.56 & 64.35 & 60.77 & 37.1 \\
\hline
\xmark & \cmark & \xmark & 68.95 & 68.34 & 64.34 & 8.2 \\
\hline
\cmark & \xmark & \xmark & 81.09 & 70.21 & 65.12 & 35.8 \\
\hline
\cmark & \cmark & \xmark & 80.97 & 71.97 & 67.83 & 30.4 \\
\hline
\cmark & \xmark & \cmark & 81.52 & 71.32 & 66.95 & 35.8 \\
\hline
\cmark & \cmark & \cmark & \textbf{82.08} & \textbf{73.48} & \textbf{69.33} & 30.4 \\
\hline
\end{tabular}
}
\label{table:ablation-single}
\end{table}

We evaluate the contribution of each component of our module on the KITTI validation set, using the single-modal detector PointPillars as a baseline.
Table \ref{table:ablation-single} reports the overall 3D AP on the KITTI validation set, aggregated w.r.t. the difficulty of the detections. The first line reports the performance of PointPillars, where we report the inference time of the RGB branch (the Faster RCNN) to have a fair comparison with the other lines. The other rows of the tables refer to computations that are added on top of the LiDAR and RGB branches. The Bbox Matching module, shown in the third row, significantly improves the metrics by reducing the FP detections, confirming our hypothesis that the precision of the RGB branch is higher due to the accurate semantic information contained in it.
It is worth noting that the RGB branch also generally exhibits a higher recall than the LiDAR branch, particularly for small or distant objects such as pedestrians, cyclists, and far-away vehicles.
The improvements observed in our ablation study (fourth row of Table \ref{table:ablation-single}) when enabling Detection Recovery confirm that many unmatched RGB detections correspond to true positives missed by LiDAR, thereby substantiating the recall advantage of the RGB branch.
In the fifth row, we show that enabling it together with the Bounding Box Matching, we further enhance the performance of Bounding Box Matching alone, confirming that the RGB branch provides a more reliable semantic information. Finally, combining all modules yields, as expected, the best overall performance. Notably, by enabling Detection Recovery only (second line), we obtain worse performance than the LiDAR branch in the easy setting, but better performance in medium and hard settings. Additionally, as expected, Detection Recovery alone is not suitable for real-time applications.

\subsubsection{Computational Complexity}
We evaluate the inference speed of LCF3D on an A100 GPU to assess real-time applicability. As shown in Tables \ref{table:multi_modal_comparison} and \ref{table:ablation-single}, PointPillars runs at 62.5 FPS and Faster R-CNN at 37 FPS. Since RGB and LiDAR branches can operate in parallel \cite{point-painting}, the overall rate is determined by the slower branch. The additional cost introduced by our three fusion modules is minimal, allowing the system to meet real-time requirements given typical LiDAR sensor rates (10–20 FPS).
Table~\ref{table:computational-complexity} further reports computational costs: LCF3D achieves an excellent trade-off between efficiency and accuracy, with most of the extra time ($\sim$ 5 ms) due to the Detection Recovery step, which depends on the number of unmatched RGB detections. Despite this, LCF3D remains faster and more memory-efficient than MVX-Net \cite{mvx-net} and BEVFusion \cite{liu2023bevfusion}, confirming that the late-cascade design yields significant accuracy gains with negligible computational penalty.

\begin{table}[t]
\centering
\caption{
Computational complexity is evaluated on KITTI, comparing single-view LCF3D with MVXNet and BEVFusion. Final inference time considers parallel execution, taking the slower branch.
}
\resizebox{0.8\textwidth}{!}{%
\begin{tabular}{|>{\centering\arraybackslash}p{5cm}|>{\centering\arraybackslash}p{3cm}|>{\centering\arraybackslash}p{3cm}|>{\centering\arraybackslash}p{3cm}|}
\hline
Module & Speed (ms) & GFLOPS & GPU Memory (MB)\\
\hhline{|=|=|=|=|}
LiDAR branch (PointPillars) & 16.04 & 62.5 & 458.17 \\
\hline
RGB branch (Faster RCNN) & 26.95 & 320.38 & 467.79 \\
\hline
Bounding Box Matching & 0.88 & 0.00 & 268.72 \\
\hline
Detection Recovery & 5.06 & 2.23 & 405.57\\
\hline
Semantic Fusion & 0.14 & 0.00 & 268.71 \\
\hhline{|=|=|=|=|}
LCF3D & \textbf{33.03} & 385.11 & \textbf{467.79} \\
\hline
MVXNet \cite{mvx-net} & 96.76 & \textbf{311.93} & 922.54 \\
\hline
BEVFusion \cite{liu2023bevfusion} & 70.37 & 411.63 & 1291.42 \\
\hline
\end{tabular}
}
\label{table:computational-complexity}
\end{table}

\subsubsection{Bounding Box Matching} \label{sec:ablation-bbox-match}

\begin{figure}[t]
\centering
\includegraphics[width=0.8\textwidth]{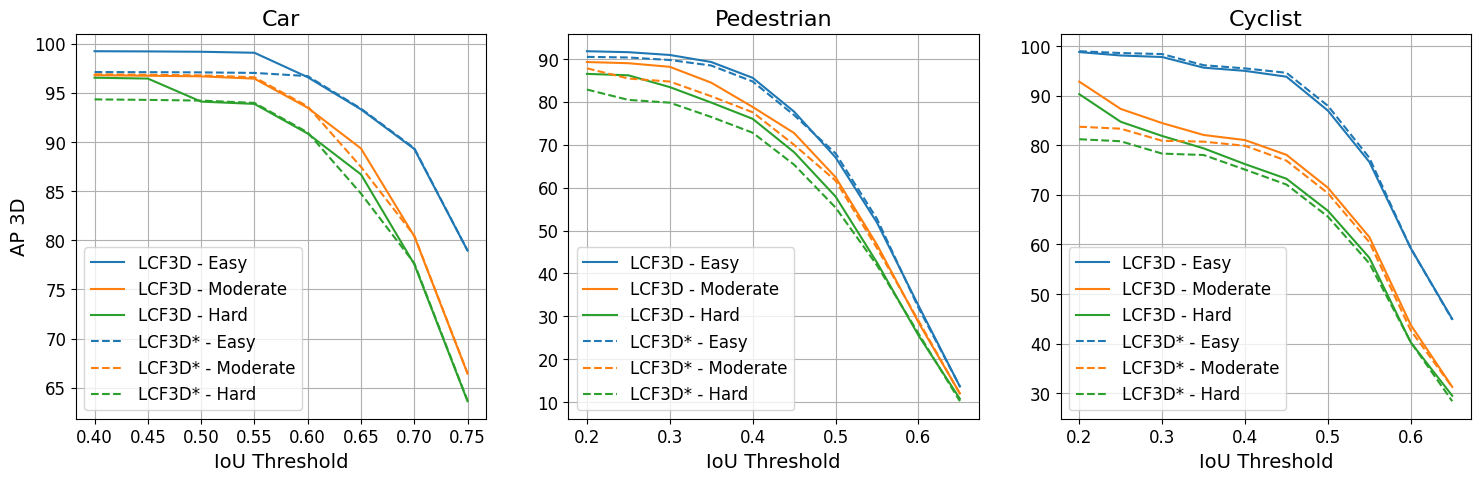}
\caption{Plots comparing the Bounding Box Matching module, by removing Detection Recovery, of LCF3D with our previous version in \cite{me}, denoted with LCF3D*.}
\label{fig:bbox-matching-comparison}
\end{figure}

We evaluate the impact of our improved Bounding Box Matching module relative to our previous work \cite{me}. To isolate its effect, we disable the Detection Recovery module and compute 3D AP across multiple IoU thresholds on the KITTI dataset. Results in Figure \ref{fig:bbox-matching-comparison} show that the new matching strategy yields higher AP at lower IoU thresholds, as removing NMS increases the likelihood of retaining True Positive LiDAR detections even with slightly inaccurate boxes. 
Since late fusion cannot correct localization errors, the final bounding box quality depends on the LiDAR branch.
Since in real-world scenarios it is often more important to detect an object, even with a slightly inaccurate bounding box, than to miss it entirely, our improved Bounding Box Matching allows us to retain True Positive objects with a low IoU with the ground truth.
Thanks to our Clustered Detections, we can preserve such objects without affecting the removal of actual FPs in the previous version of the Bounding Box Matching.

\subsubsection{2D Object Detection vs Instance Segmentation}

\begin{table}[t]
\centering
\caption{The effect of instance segmentation masks in the Frustum Localizer performance (Cascade Fusion), using DetectorRS as RGB detector. The FPS are reported for the Detection Recovery module only.}
\resizebox{0.75\textwidth}{!}{%
\begin{tabular}{|c||>{\centering\arraybackslash}p{1.4cm}|>{\centering\arraybackslash}p{1.4cm}||>{\centering\arraybackslash}p{1.4cm}|>{\centering\arraybackslash}p{1.4cm}|>{\centering\arraybackslash}p{1.4cm}||>{\centering\arraybackslash}p{1.4cm}|}
\hline
Frustum Proposals & mAP $\uparrow$ & NDS $\uparrow$ & Pedestr. $\uparrow$ & Motor. $\uparrow$ & Bicycle $\uparrow$ & FPS $\uparrow$ \\
\hline
Bbox & 0.265 & 0.283 & 0.447 & 0.305 & 0.347 & 5.26 \\
\hline
Bbox + Mask & 0.295 & 0.306 & 0.461 & 0.365 & \textbf{0.382} & 4.54 \\
\hline
Mask & \textbf{0.316} & \textbf{0.333} & \textbf{0.531} & \textbf{0.379} & 0.377 & \textbf{6.67} \\
\hline
\end{tabular}
}
\label{table:cascade-fusion-ablation}
\end{table}

In Table \ref{table:cascade-fusion-ablation}, we compare the three approaches for extracting Frustum Proposals shown in Figure \ref{fig:frustums}. Namely, i) \emph{Bbox} extracts proposals from 2D bounding boxes, ii) \emph{Mask} uses instance segmentation masks, and iii) \emph{Bbox+Mask} adds the mask as an additional channel to the bounding box input. We report the performance of a vanilla cascade fusion setup where the LiDAR branch and Bounding Box Matching module are removed, i.e. all the RGB 2D detections are used as input for the Detection Recovery. We report global mAP, NDS, and per-class mAP for Pedestrians, Motorcycles, and Bicycles. Results show that adding the mask channel improves metrics of \emph{Bbox} but increases computational cost. Finally, \emph{Mask} obtains the best performance as it selects only the points that project inside the instance mask, and has a lower computational overhead, as fewer points are kept in Frustum Proposals. 
These results confirm that LiDAR branch and late fusion are necessary, as cascade fusion alone is insufficient for real-time performance, and LCF3D outperforms it using the same RGB detector.

\subsection{Domain Generalization analysis}

\begin{table}[t]
\centering
\caption{Performance of our method under domain shifts. We report the $NDS^{\hat{*}}$ defined as in \eqref{eq:nds_hat}.}
\resizebox{0.75\textwidth}{!}{%
\begin{tabular}{|c|c|>{\centering\arraybackslash}p{1.4cm}|>{\centering\arraybackslash}p{1.4cm}|>{\centering\arraybackslash}p{1.4cm}|>{\centering\arraybackslash}p{1.4cm}|>{\centering\arraybackslash}p{1.4cm}|}
\hline
Task $(Train \rightarrow Test)$ & Model & $NDS^{\hat{*}} \uparrow$ & $mAP \uparrow$ & $mATE \downarrow$ & $mASE \downarrow$ & $mAOE \downarrow$\\
\hhline{|=|=|=|=|=|=|=|}
\multirow{5}{*}{$Kitti \rightarrow Kitti$} & PointPillars \cite{pointpillars} & 0.722 & 0.696 & 0.101 & 0.214 & 0.443\\
\cline{2-7}
& MVXNet \cite{mvx-net} & 0.714 & 0.712 & 0.111 & \textbf{0.204} & 0.537\\
\cline{2-7}
& BEVFusion \cite{liu2023bevfusion} & 0.804 & 0.771 & \textbf{0.099} & 0.227 & \textbf{0.161} \\
\cline{2-7}
& Ours & \textbf{0.821} & \textbf{0.908} & 0.101 & 0.222 & 0.475 \\
\hline
\multirow{5}{*}{$Kitti \rightarrow nuScenes$} & PointPillars & 0.152 & 0.059 & 0.731 & 0.739 & 0.798 \\
\cline{2-7}
& MVXNet & 0.127 & 0.025 & 0.768 & 0.752 & \textbf{0.791} \\
\cline{2-7}
& BEVFusion & \textbf{0.243} & 0.091 & \textbf{0.408} & \textbf{0.406} & 1.232 \\
\cline{2-7}
& Ours & 0.234 & \textbf{0.157} & 0.516 & 0.552 & 1.067 \\
\hhline{|=|=|=|=|=|=|=|}
\multirow{5}{*}{$nuScenes \rightarrow nuScenes$} & PointPillars & 0.577 & 0.427 & 0.267 & 0.245 & 0.309 \\
\cline{2-7}
& MVXNet & 0.586 & 0.452 & 0.233 & 0.251 & 0.356\\
\cline{2-7}
& BEVFusion & \textbf{0.685} & \textbf{0.612} & \textbf{0.181} & \textbf{0.242} & \textbf{0.306} \\
\cline{2-7}
& Ours & 0.539 & 0.459 & 0.305 & 0.310 & 0.528 \\
\hline
\multirow{5}{*}{$nuScenes \rightarrow Kitti$} & PointPillars & 0.455 & 0.414 & 0.189 & \textbf{0.325} & 1.048 \\
\cline{2-7}
& MVXNet & 0.449 & 0.379 & 0.178 & 0.340 & 0.926 \\
\cline{2-7}
& BEVFusion & 0.525 & 0.474 & \textbf{0.169} & 0.359 & \textbf{0.743} \\
\cline{2-7}
& Ours & \textbf{0.545} & \textbf{0.613} & 0.198 & 0.371 & 1.094 \\
\hline
\end{tabular}
}
\label{table:ablation-domain-shifts}
\end{table}

\begin{figure}[t]
    \centering
    \includegraphics[width=0.8\linewidth]{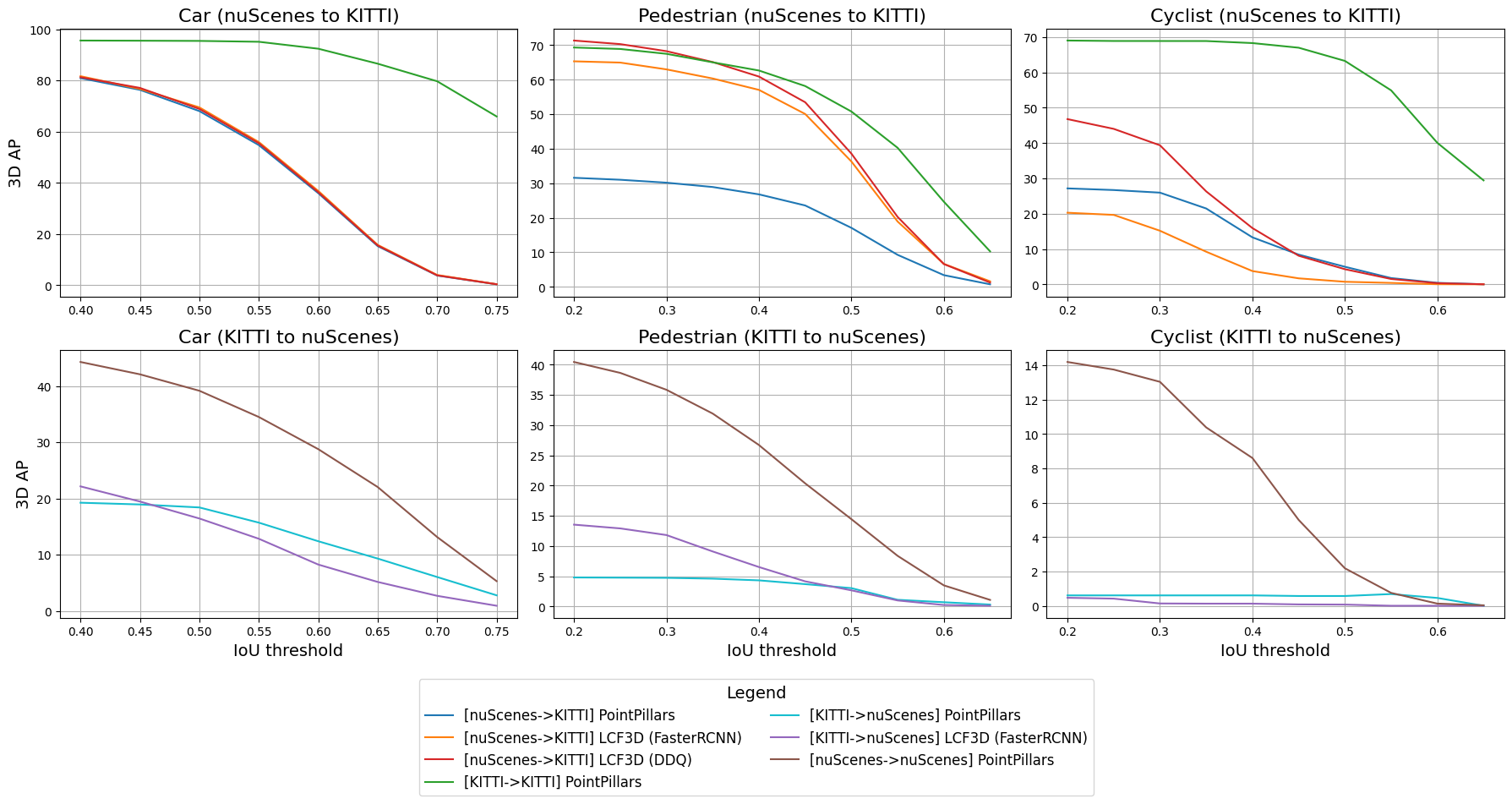}
    \caption{The 3D Average Precision, as the IoU threshold varies, under distribution shifts.}
    \label{fig:domain-shift}
\end{figure}

We evaluate the performance of LCF3D when tested on a different dataset from the one it was trained on. For models trained on nuScenes, sweeps are used to augment the Point Cloud and usually an additional channel, representing the timestamp difference w.r.t the current frame, is added to each point. Since KITTI lacks sweeps, to test nuScenes models on KITTI, we perform inference on a single frame and set this channel to zero in each 3D point. For MVXNet and BEVFusion, we use the open-source implementations from MMDetection3D. We train MVXNet on KITTI for 40 epochs by using the suggested parameters by the framework, and on nuScenes for 24 epochs using ResNet-50 as backbone for the image modality and SECOND as voxel encoder. For BEVFusion, we use the pre-trained nuScenes model by MMDetection3D, and train the KITTI model by pre-training the LiDAR backbone for 3D Object Detection and fine-tuning it together with the RGB modality.

Table \ref{table:ablation-domain-shifts} shows nuScenes metrics. While BEVFusion outperforms across all metrics on nuScenes, under domain shift (e.g., nuScenes $\to$ KITTI) our method achieves higher mAP, despite BEVFusion having lower mATE, mASE, and mAOE for matched true positives.
Since mAP reflects the ability to correctly detect objects (penalizing both FPs and FNs), this result indicates that our approach is more robust in maintaining correct detections under domain shift, even though it cannot improve the precision of the bounding boxes themselves. The reason is that our framework leverages pre-estimated LiDAR detections: the geometry of each bounding box is bounded by the LiDAR detector’s performance, but the Bounding Box Matching procedure reduces the number of spurious detections, leading to fewer FPs, and the Detection Recovery can find missed objects, leading to fewer FNs.
This robustness is important in real-time scenarios, where missing or hallucinating objects are more critical than slight inaccuracies in box geometry. Additionally, using a more generalizable LiDAR detector could directly improve mATE, mASE, and mAOE.

Figure \ref{fig:domain-shift} shows the 3D AP w.r.t. IoU thresholds. On the KITTI dataset, LCF3D almost recovers the performance of PointPillars trained directly on KITTI, by fusing a PointPillars model trained on nuScenes with a 2D RGB detector (either DDQ or Faster R-CNN) trained on nuScenes and nuImages. However, for Cars, we are unable to surpass PointPillars. This can be explained by the 3D AP values of PointPillars trained on nuScenes: the 3D AP  exceeds $80\%$ at an IoU threshold of 0.4, indicating that the model correctly detects many True Positives, albeit with inaccurate bounding boxes. 
Conversely, for models trained on KITTI and tested on nuScenes, a large performance gap remains even at lower IoUs, mainly due to the higher sparsity of nuScenes point clouds and the fact that KITTI models are trained on single-frame data without leveraging multiple sweeps. Please note that Cyclist performance is affected by differences in dataset definitions: KITTI does not annotate bicycles without riders, unlike nuScenes.

\subsection{Limitations and Qualitative Results}

\begin{figure}[t]
\centering
\includegraphics[width=0.9\textwidth]{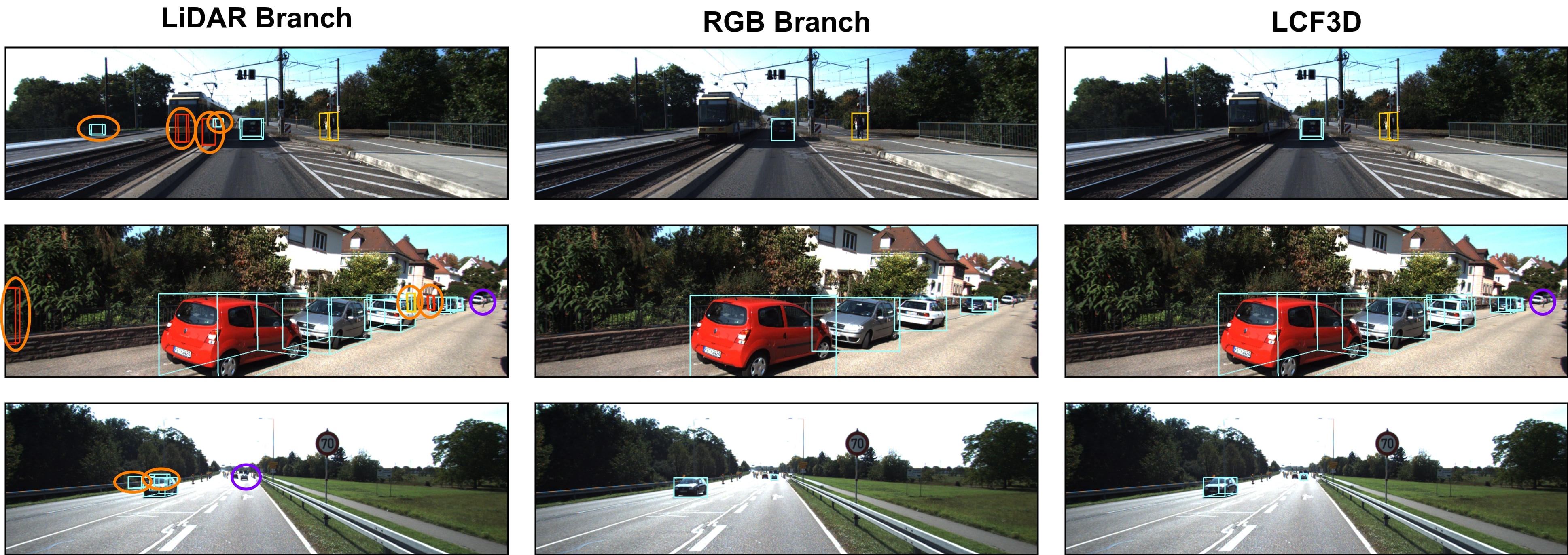}
\caption{
Qualitative results on KITTI show object classes by bounding box color: red for pedestrians, yellow for bicycles, and cyan for cars; orange circles indicate FPs, purple circles FNs.
GB (Faster R-CNN) improves precision by removing LiDAR FPs and can recover missed objects (the car in the last frame) though recovery fails if both branches miss them (second row).}
\label{fig:kitti-visualization}
\end{figure}

\begin{figure}[t]
\centering
\includegraphics[width=0.9\textwidth]{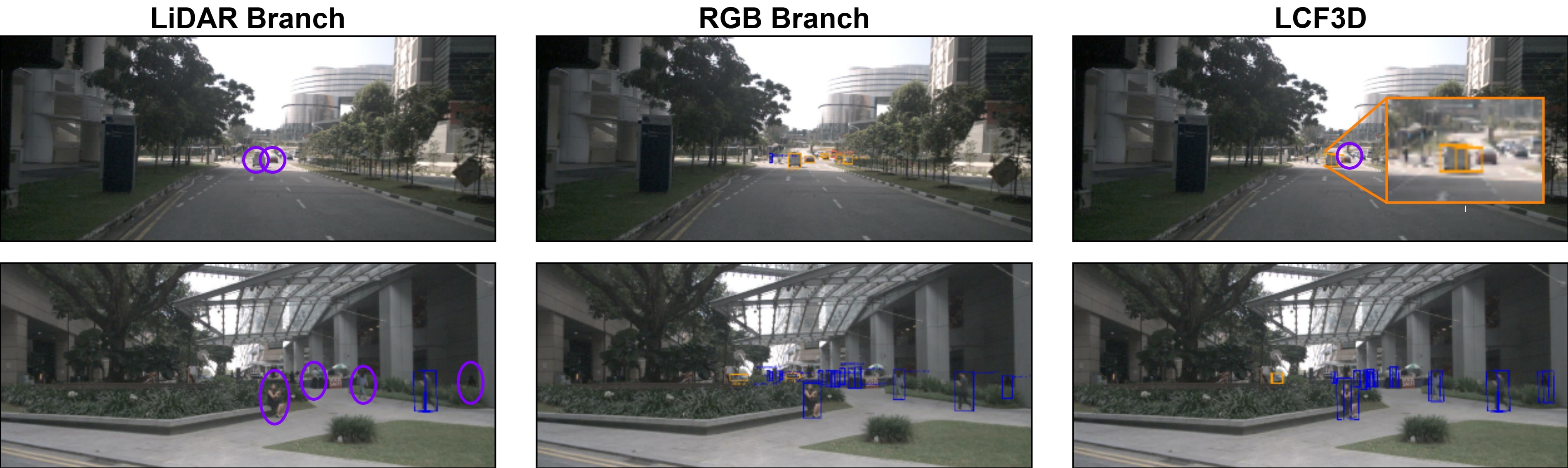}
\caption{Qualitative results on the nuScenes dataset.
Sparse LiDAR point clouds cause more FNs for distant objects. Our method recovers many, but limited points hinder full recovery and orientation estimation.
}
\label{fig:nuscenes-visualization}
\end{figure}

Figures~\ref{fig:kitti-visualization} and \ref{fig:nuscenes-visualization} provide qualitative examples on KITTI and nuScenes respectively, highlighting the complementary strengths of LiDAR and RGB. On KITTI, the RGB branch improves precision by suppressing LiDAR FPs and recovers distant or small objects missed by LiDAR, though recovery fails when both branches miss an object (second row). On nuScenes, sparser LiDAR point clouds cause more FNs, many of which are recovered by our method, but some frustums lack sufficient points for reliable 3D localization. Orientation estimation from frustums is also challenging with few points (top-right image). 
Overall, qualitative results show that LCF3D reduces FPs via late fusion and recovers FNs through cascade fusion, achieving strong cross-dataset performance, though some limitations remain. LCF3D requires both modalities; if one branch fails systematically, recovery is impossible. Additionally, recovery quality depends on LiDAR point density within frustums, with sparse scenes potentially causing inaccurate localization or orientation.

\section{Conclusions}

We have proposed a hybrid late-cascade fusion approach that exploits a 3D LiDAR detector, a 2D RGB detector and the geometrical constraints of the scene.
Our solution increases the performance of single-modal detectors, especially for more challenging classes like Cyclists and Pedestrians and is completely independent of the underlying single-modal detectors, allowing flexible solutions including the usage of pre-trained state-of-the-art models. 
Computationally, LCF3D introduces minimal overhead and offers a strong balance of latency, memory, and accuracy compared to other multimodal approaches, making it suitable for real-world autonomous driving. Limitations include reliance on both modalities and sensitivity to sparse point clouds, which can affect 3D localization and orientation. Future work will explore more robust frustum-based localization and alternative recovery mechanisms using RGB data to compensate for missing LiDAR information.

\section*{Acknowledgements}
This paper is supported by  FAIR  (NextGenerationEU program, PNRR-PE-AI scheme, M4C2, Investment 1.3, Line on Artificial Intelligence) and by GEOPRIDE ID: 2022245ZYB, CUP: D53D23008370001 (PRIN 2022 M4.C2.1.1 Investment).
Model training and testing were possible thanks to the HPC grant from by the Ministry of Education, Youth and Sports of the Czech Republic through the e-INFRA CZ (ID:90254)





\bibliographystyle{elsarticle-num} 
\bibliography{bibliography.bib}






\end{document}